\documentclass[pdflatex,sn-mathphys,Numbered,iicol]{sn-jnl}

\usepackage{graphicx}%
\usepackage{multirow}%
\usepackage{amsmath,amssymb,amsfonts}%
\usepackage{amsthm}%
\usepackage{mathrsfs}%
\usepackage[title]{appendix}%
\usepackage{xcolor}%
\usepackage{textcomp}%
\usepackage{manyfoot}%
\usepackage{booktabs}%
\usepackage{algorithm}%
\usepackage{algorithmicx}%
\usepackage{algpseudocode}%
\usepackage{listings}%
\usepackage{scrextend}
\usepackage{bigfoot}
\usepackage[justification=justified,font=small]{caption}

\begin{document}

\title[Parameter Estimation using Reinforcement Learning Causal Curiosity: Limits and Challenges]{Parameter Estimation using Reinforcement Learning Causal Curiosity: Limits and Challenges}

\author*[1]{\fnm{Miguel} \sur{Arana-Catania}}\email{humd0244@ox.ac.uk}
\author[2]{\fnm{Weisi} \sur{Guo}}

\affil[1]{\orgdiv{Digital Scholarship at Oxford}, \orgname{University of Oxford}, \orgaddress{\country{UK}}}
\affil[2]{\orgdiv{Faculty of Engineering and Applied Sciences}, \orgname{Cranfield University}, \orgaddress{\country{UK}}}

\abstract{Causal understanding is important in many disciplines of science and engineering, where we seek to understand how different factors in the system causally affect an experiment or situation and pave a pathway towards creating effective or optimising existing models. Examples of use cases are autonomous exploration and modelling of unknown environments or assessing key variables in optimising large complex systems. In this paper, we analyse a Reinforcement Learning approach called Causal Curiosity, which aims to estimate as accurately and efficiently as possible, without directly measuring them, the value of factors that causally determine the dynamics of a system. Whilst the idea presents a pathway forward, measurement accuracy is the foundation of methodology effectiveness. Focusing on the current causal curiosity's robotic manipulator, we present for the first time a measurement accuracy analysis of the future potentials and current limitations of this technique and an analysis of its sensitivity and confounding factor disentanglement capability - crucial for causal analysis. As a result of our work, we promote proposals for an improved and efficient design of Causal Curiosity methods to be applied to real-world complex scenarios.}

\keywords{reinforcement learning, dynamical systems, causal analysis, sensitivity analysis, parameter estimation}



\maketitle

\section{Introduction}\label{sec:intro}

In the study of dynamical systems, it is a common need to estimate the value of parameters that determine their interactions and dynamics. These may be unknown, for example, because they belong to an unexplored environment (e.g. the weight of a lunar rock to be picked up by a rover on a space mission, or the magnitude and direction of the wind field for drone operations), or because our sensors and system model presents errors or uncertainties in the value of its parameters (e.g. the distance between the wheels of a mobile autonomous robot may vary after interacting with a rough terrain). In the case of autonomous systems, such as the examples above, it might not be feasible to access the system to perform measurements, or they might not have the appropriate tools because of design reasons. In these situations, it is crucial to have indirect strategies that allow estimating their value as accurately as possible. Here, we focus on a general and established experimental case of a robotic manipulator interacting with physical objects with unknown parameters, and the use of machine learning to estimate these parameters. 

In this work, we approach this problem from the point of view of causal analysis. The enormous development and application of machine learning techniques witnessed in recent years are allowing to discover some of their main limitations. One of these limitations is the fact that machine learning techniques are usually designed to find correlations in the data during the learning process but not to find causation or apply proper causal reasoning \cite{scholkopf2022causality, scholkopf2021toward,locatello2019challenging,suter2019robustly,scholkopf2012causal,kilbertus2018generalization,lu2018deconfounding}. This results in models that are unable to correctly solve tasks that are nevertheless solvable with causal reasoning, e.g. avoiding spurious relationships due to data correlations or reasoning using the correct causal relations. Examples can be seen in fields such as computer vision \cite{singla2021salient,beery2018recognition}, reinforcement learning \cite{wang2022causal,de2019causal,ortega2021shaking}, natural language processing \cite{niu2021counterfactual,alzantot2018generating}, or graph analysis \cite{chen2022invariance,feng2021should}. As a result of this, an increasing number of works are trying to bridge this gap by combining causal analysis and machine learning.

From a causal point of view, the parameters to be estimated in our system are denoted as causal factors, since their values causally determine the dynamics of the system. The objective of the Causal Curiosity technique is to return strategies to interact with the unknown objects that are most effective in estimating these causal factors. In the language of causal analysis, we look for interventions that allow us to estimate the Causal Directed Acyclic Graphs (DAG) of our system \cite{pearl2000models,glymour2016causal,peters2017elements,spirtes2000causation}. In the case analysed in this work, this translates into identifying which are the most appropriate movements of our robotic manipulator to estimate the unknown parameters of the object (e.g., its mass or friction coefficient).

Bringing together the use of a machine learning technique in the estimation of different parameters and using a causal approach, we work here with the recent proposal of Causal Curiosity \cite{sontakke2021causal}. This approach has been designed to determine the fundamental parameters of an unknown system in a systematic and organised way, following a strategy of scientific experimentation on these factors. It shows an enormous potential since it proposes a robust, explainable methodology, anchored in a causal analysis framework that thus allows going beyond the previously mentioned limitation of common methodologies in Reinforcement Learning (RL). 

This method presents fundamental advantages in comparison to other methodologies. Previous works on parameter estimation with robotic manipulators show the possibility of estimating, for example, mass, stiffness, inertia, hardness, shape, texture, friction, etc \cite{thompson2022identification, mkc2018inertia, murali2018learning, yuan2017shape, yi2016active, kaboli2017tactile, yu2017preparing, murooka2017feasibility, mavrakis2020estimation}. However, these methodologies are not generalisable, in the sense that they can only estimate certain types of parameters and need specific formulations for each type of parameter (e.g. in general a method for estimating the inertia of an object cannot be used to estimate its friction coefficient). This is partly due to the fact that traditional methodologies commonly use knowledge of the equations that define the dynamics of the system (e.g. using a linear approximation of the equations to obtain the relationship between parameters). In our case, it is sufficient to simulate the dynamics of the system, without the need for explicit knowledge of the system equations or to perform operations on them to understand the effect of the parameters to be estimated on the observations of the system. Thus, the method is generalisable, in the way that it can be applied to any type of system parameter, and do so without main methodological changes in the implementation regardless of the parameter we consider.
A second fundamental advantage is the ability to obtain an explainable methodology. As we will see later, the methodology maximises a human-explainable reward. Furthermore, the formulation of the problem is done in a causal framework, where the parameters affecting the system and the causal relationship between them are clearly defined, which is fundamental for its explainability.
As a third fundamental advantage, the causal approach is an upgrade over non-causal machine learning methodologies. The latter do not consider in any way the causal relationship of the parameters on which they act and therefore fail in the case of confounding relationships between the variables on which they are trained. In our case, the analysis of the system always takes into account the causal relationships and therefore identifies beforehand the possible limitations that may exist, and the impact that these may have on the estimation of the parameters, as we will see especially in the last experiments carried out in this work.

Although there are many advantages to this methodology, the original proposal of this approach \cite{sontakke2021causal} leaves unanswered how much of this potential can be realised in practice when dealing with complex systems.

The main contribution of this work is the analysis of this technique, in this case examined in the use case of a robotic manipulator, by confronting it for the first time with limiting situations in the complexity of the system to be explored or in the sensitivity sought in the parameter estimation, with the aim of understanding its limits and challenges. As an additional contribution, we also propose changes in the methodology to increase its effectiveness. 

The implementation of the robotic manipulator is done using the same simulation framework as the article of the original proposal, CausalWorld\footnote{\url{https://github.com/rr-learning/CausalWorld}}, which is a framework specifically designed to work with causal approaches.

Our results confirm the enormous potential of the methodology and point out scenarios where difficulties are encountered and new lines of future work are needed to advance this methodology. Among other results, regarding its potential, we show the possibility of refining the accuracy of parameter identification by several orders of magnitude while increasing the robustness of the analysis by about 15\% in the case of our proposed methodology. In relation to its limitations, we show the impossibility of solving some complex experimental situations studied for the first time here with multiple parameters varying simultaneously. We also analyse, with positive results, cases in which the factors present causal relationships not only with respect to the general evolution of the system dynamics, but also between them, and in particular we study a case with confounding variables.

We have divided our analysis into five strands, in order to obtain a result as comprehensive as possible. We focus our research questions (RQ) on developing the measurement accuracy analysis of the following areas that enable improved causal analysis:

\begin{itemize}
\item RQ1. Accuracy in the estimation of different causal factors.

\item RQ2. Granularity in the determination of causal factors.

\item RQ3. Gap size effect in the estimation of causal factors.

\item RQ4. Multiple causal factors determination.

\item RQ5. Causally related and confounding causal factors estimation.
\end{itemize}

Additionally, we evaluate the use of Proximal Policy Optimization (PPO) \cite{schulman2017proximal} as a causal factor estimation method and compare it to the Cross-Entropy Method (CEM) optimised Model Predictive Control Planner \cite{de2005tutorial,camacho2013model} proposed by the authors of Causal Curiosity.

\section{Related Work}\label{sec:relwork}

The fields of Causal Analysis  \cite{pearl2000models,glymour2016causal,peters2017elements,spirtes2000causation} and Reinforcement Learning are increasingly being combined as a way of overcoming the inevitable limitations of the traditional correlational approach to RL that give rise to problems of generalisability, robustness and inefficiency, among others. This is leading to the development of the emerging fields of Causal Machine Learning and Causal RL  
\cite{kaddour2022causal, scholkopf2022causality, scholkopf2021toward, bareinboimcrlonline, zeng2023survey, grimbly2021causal, weichwald2022learning,bareinboim2015bandits,gershman2017reinforcement,dasgupta2019causal}.

In this latter field, several works introduce causal concepts to try to improve the inefficiency in exploring the state space, one of the main problems in RL (traditionally not related to causality, see \citet{amin2021survey}). For example, \citet{peng2022causality} propose a causality-driven hierarchical RL framework using causality-driven exploration and subgoal structures to improve the exploration phase in challenging tasks with sparse rewards, \citet{rezende2020causally} work with partial models as a way to avoid learning the full model in intractable high-dimensional spaces, and use causality to correctly connect such partial models, \citet{pitis2020counterfactual} introduce local causal models, which are induced from a global causal model by conditioning on a subset of the state space, to improve the sample efficiency and performance of RL systems that involve sub-processes, 
\citet{seitzer2021causal} improve exploration by using the local causal graph to guide the learning based on what can be influenced in each state, \citet{molina2020causal} use knowledge from causal models to reduce the exploration space of RL models.

A key element in our approach is the concept of curiosity. The field of RL has long considered intrinsic motivation approaches and curiosity-driven learning as a way to solve complex tasks and deal with the exploration-exploitation dilemma  \cite{pathak2017curiosity,burda2018large,schmidhuber1991curious,chentanez2004intrinsically,lehman2008exploiting,oudeyer2009intrinsic,sun2011planning,still2012information,  baldassarre2013intrinsically, baranes2013active,  barto2013intrinsic,stadie2015incentivizing,mohamed2015variational,houthooft2016vime,osband2016deep,forestier2017intrinsically,tang2017exploration, colas2018gep,laversanne2018curiosity, oudeyer2018computational, guo2022byol}.

However, Causal Curiosity is not only linked to efficiently exploring the state space as in most of the previous works but it is ultimately related to the goal of correctly identifying causal factors through experimentation. Other works with a focus on intervention and experimentation to discover causal relationships are the following: \citet{gasse2021causal} propose a model-based Causal RL method where the information obtained from the interactions of the agent with the environment are combined with observational offline data from another agents interactions, \citet{dasgupta2019causal} use meta-learning to train agents in tasks depending on a causal structure, and show how the agents become capable of performing experiments that can be seen as causal reasoning, \citet{ke2021systematic} study causal induction in model-based reinforcement learning, \citet{nair2019causal} propose techniques for causal induction from raw visual observations and causal graph encoding, \citet{ding2022generalizing} propose a method that alternates between performing interventions to estimate the causal graph and using the graph to learn generalisable models, \citet{thomas2017independently} work on objective functions to obtain a disentangled representation through interaction with the environment, \cite{burgess2018understanding, kim2018disentangling, chen2018isolating} work in obtaining disentangled representations using variational autoencoders, \citet{volodin2020resolving} intervene in an unknown environment to solve spurious correlations.

\section{Methodology}\label{sec:method}

In this section, we introduce the conceptual framework of causal analysis used in this paper \cite{pearl2000models,glymour2016causal,peters2017elements,spirtes2000causation} and summarise the main theoretical elements of the Causal Curiosity approach \cite{sontakke2021causal} and the details of the specific methodology applied in this paper.

\subsection{Simulation framework and main parameters}

The framework used in this research is CausalWorld\footnote{\label{causalworldfootnote}\url{https://github.com/rr-learning/CausalWorld}}. This is an open-source simulation framework and benchmark for causal structure and transfer learning in a robotic manipulation environment. This framework has been designed to allow an easy analysis of a dynamic system from a causal point of view, enabling in addition the application of reinforcement learning methodology.

\begin{figure}[ht]
    \centering
    \includegraphics[width=1.00\linewidth]{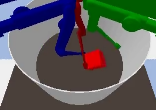}
    \caption{CausalWorld simulation}
    \label{fig:robot}
\end{figure}

CausalWorld uses the Bullet physics engine \cite{coumans2013bullet} to simulate the open-source TriFinger robot platform \cite{wuthrich2021trifinger}. This framework allows the modification of different physical parameters that causally determine the evolution of the system's interaction and dynamics. A screenshot of the simulation can be seen next in Figure \ref{fig:robot} and videos of the robot in operation can be found in the framework site\footref{causalworldfootnote}.

In this work, the robot manipulator interacts with a single object, a cube whose dynamic is defined by the following causal factors: \textbf{mass}, \textbf{size}, \textbf{lateral friction}, \textbf{spinning friction}, and \textbf{gravity}. The first four factors affect the object, while the last one affects both the object and the robot.

In Table \ref{tab:factors_values} we present the initial value of these parameters and the maximum range of variation that we explore.

\begin{table}[ht]
    \centering
    \begin{tabular}{c|cc}
    \toprule
    \textbf{Factor} & \textbf{Value} & \textbf{Range} \\
    \midrule
    \textbf{Mass} & 0.25 &  [0.01,0.5] \\
    \textbf{Size}  & 0.075 & [0.05,0.1] \\
    \textbf{Lat. Frict.}  & 1 & [0.1,1.0]\\
    \textbf{Spin. Frict.}  & 0.001 & [0.001,1.0] \\
    \textbf{Gravity}  & -9.81 & [-1.0,-11.5] \\
    \bottomrule
    \end{tabular}
\caption{Causal factors values and ranges.}
    \label{tab:factors_values}    
\end{table}

\subsection{Causal framework}

From the point of view of causal analysis \cite{pearl2000models,glymour2016causal,peters2017elements,spirtes2000causation}, the variables $x_i$ that define the system are related to each other by a series of functions $f_i$ called structural assignments: 

\begin{equation}
x_i \equiv f_i (\epsilon_i ; pa_i)
\end{equation}

These functions determine how each variable changes influenced by the other variables (called parents, and denoted by $pa_i$), and by independent exogenous noise variables $\epsilon_i$. 

The set of structural assignments is called the Structural Causal Model (SCM):

\begin{equation}
S = \left\{ f_i \right\}^N_{i=1}
\end{equation}

The SCM is a complete description of the system from a causal point of view and entails its observational, interventional and counterfactual distributions.

As will be seen in more detail below, in our case these structural assignments are the dynamical equations of the system that connect the parameters defining the object with which we will interact (mass, size, etc.) with the observations during the experiments (represented as the trajectories of the object).

Commonly, several principles are generally assumed \cite{peters2017elements}. Firstly, the Principle of Independent Mechanisms, which from a probabilistic point of view indicates that the conditional distribution of each variable given its causes (i.e. its mechanism) does not inform or influence the other conditional distributions.

This principle allows us to disentangle the chain rule of probability,

\begin{equation}
p\left(x_1, ..., x_n \right) = \prod_{i} p\left(x_i \mid x_1,...,x_{i-1} \right) \end{equation}

and express the probability of the system as the product of probabilities of variables that depend only on their parents through the Markov factorisation property:

\begin{equation}
p\left(x_1, ..., x_n \right) = \prod_{i} p\left(x_i \mid pa_i \right) 
\end{equation}

\begin{figure}[ht]
    \centering
    \includegraphics[width=1.00\linewidth]{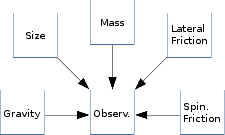}
    \caption{Causal DAG}
    \label{fig:dagsall}
\end{figure}

We also assume that the variables are related in a non-cyclic manner which implies that every SCM induces a Causal Directed Acyclic Graph (DAG) as depicted in  Figure \ref{fig:dagsall}. In this figure, we represent the DAG corresponding to the system studied in this article, with the Observation node representing the trajectory of the object, which is affected, as indicated by the arrows, by the different parameters mentioned above.

\subsection{System dynamics, causal factors and reinforcement learning}

In addition to the SCM representing the causal relationships of the system, from a dynamic point of view the system is defined by a Partially Observable Markov Decision Process (POMDP) \cite{sutton2018reinforcement}. This is a tuple $\mathcal{M} = (\mathcal{S},\mathcal{A},\mathcal{O},P,E,R)$, where $\mathcal{S}$ is the state space, $\mathcal{A}$ is the action space, $\mathcal{O}$ is the observation space, $P$ describes the dynamics of the system through a conditional probability distribution  $P\left( \textbf{s}_{t+1} \mid \textbf{s}_{t},\textbf{a}_{t}\right)$ with $\textbf{s} \in \mathcal{S}$ and $\textbf{a} \in \mathcal{A}$, $E$ is the emission function defining the distribution $E\left( \textbf{o}_{t} \mid \textbf{s}_{t}\right)$  with $\textbf{o} \in \mathcal{O}$, and $R$ is the reward function. In this case, the states and observations are determined by the robot agent's actions and the previous causal factors (mass, size, etc.). The \textbf{observations} represent the object trajectory, and the \textbf{actions} the movements of the robot manipulator.

Each simulation is defined by an environment, for which we can define variations in which a subset of factors modify its value (e.g. we will consider various values for the mass of the object). In causal language, we represent setting each parameter to a specific value by using the $do(\cdot)$ operator, which indicates an intervention on that parameter. We recall here the difference between an observational probability distribution of an $\textbf{o}$ variable $P \left( \textbf{o} \mid \textbf{d}'\right)$  in which we observe the distribution of $\textbf{o}$ given that we observe the variable $\textbf{d}$ for a certain fixed value $\textbf{d}'$ (i.e. we limit our observations to only those with $\textbf{d}=\textbf{d}'$) and the interventional probability distribution $P \left( \textbf{o} \mid do(\textbf{d}')\right)$  representing the distribution of $\textbf{o}$ if we set $\textbf{d}=\textbf{d}'$. These two distributions are not equivalent.

The causal factors $\textbf{h}$ that we want to estimate are parameters such that by applying a certain sequence of actions on the different variations of an environment, the probability of the observations vary in such a way that the trajectories of the objects are organised in distinguishable disjoint sets according to the values of the causal factors:

\begin{equation}
p\left(\textbf{o} \mid do(\textbf{h}=\textbf{h}'),\textbf{a} \right) \neq p\left(\textbf{o} \mid do(\textbf{h}=\textbf{h}''),\textbf{a} \right) 
\end{equation}

E.g., if we set the mass of the object to values with very low and very high mass, we expect these two groups of environments to produce object trajectories that can be organised in two disjoint sets such as large and small displacement trajectories.

The sequence of actions that cluster the trajectories in a distinguishable way is not known in advance, and thus we need strategies to search for these sequences.

\begin{figure}[ht]
    \centering
    \includegraphics[width=1.00\linewidth]{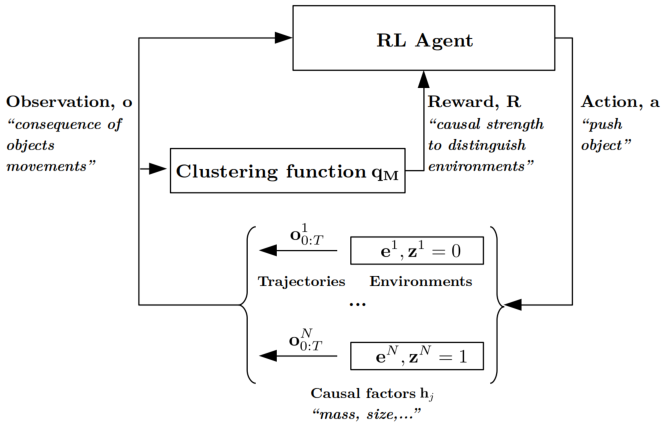}
    \caption{RL interaction}
    \label{fig:causalrl}
\end{figure}

A representation of the RL interaction can be seen in Figure \ref{fig:causalrl}.

To clarify all this, we explain next in more detail the case analysed in this work. In each one of our experiments, we consider a set of $N$ environments $\textbf{e}^i$, each of them determined by a set of $K$ causal factors $\textbf{h}_j$. In each interaction, we determine the value of only one of these causal factors, which presents a bimodal distribution. The interaction of the robot with the object aims to produce observations in the object's trajectory $\textbf{o}^i_{0:T}$ for $T$ timesteps that can be identified with two clusters corresponding to the two modes of the causal factor value. For example, the objects may have values for the mass belonging to two classes: light objects and heavy objects; a certain thrust produces two distinguishable behaviours (whose trajectories can therefore be clustered into two groups): in the light case the object falls on one of its sides, and in the heavy case the object oscillates and return to its initial position. The method explores different movements of the robot until it finds that particular thrust that is able to clearly distinguish between the two groups of environments.

In all experiments, all ranges of values are partitioned in such a way that the two ranges have the same size which is also equal to the difference between the two ranges. The only exception is presented in Section \ref{subsec:gap} where this alternative is explicitly investigated. 

In the initial Causal Curiosity proposal, all other causal factors were held constant in each individual experiment. In our case, in several experiments (see sections \ref{subsec:multiple} and \ref{subsec:confounding}) other factors vary simultaneously, to study how the combined effect of these variations affects the estimation of the main causal factor.

In our experiments, we investigate in a practical way the limits and challenges in finding the appropriate sequences of actions that distinguish between the different factors (i.e. the actions that estimate the parameters) and how distinguishable these factors are (i.e. what is the accuracy of the estimation).

\subsection{Parameter distinguishability and rewards}
The next element to explain in our methodology is how we quantify whether the trajectories are distinguishable in terms of the causal factors. For this, we need to define a reward to quantify it, which will guide the search for optimal actions.

In the initial proposal, it was chosen to minimise the Minimum Description Length as a substitute for the Kolmogorov Complexity \cite{rissanen1978modeling,grunwald2007minimum}. In practice, and considering the bimodal character of the causal factor, this was implemented as a trajectory clustering problem using a Silhouette Score $S$ \cite{rousseeuw1987silhouettes} with a distance in the trajectory space defined by Soft Dynamic Time Warping \cite{cuturi2017soft}. In our case, we keep this as part of our reward. However, since we are exploring the limits of this methodology, we face environments where it is not possible to correctly identify all the environments. This implies the need to introduce an additional value that quantifies the correct identification of the scenarios. We implement this value by means of an F1 classification score $C$. Hence, our reward is given by the sum of these two elements:

\begin{equation}
R = C\left(q_M\left(\textbf{o}^i_{0:T}\right),\textbf{z}^i\right) + k \cdot S\left(\textbf{o}^i_{0:T}\right)
\end{equation}
where $q_M$ is the clustering function learnt from the trajectories, $\textbf{z}^i\in\{0,1\}$ is the true cluster membership identifying the two groups of environments, and $k$ is a parameter defined to weight the importance of the two reward objectives, in our case set to 0.1 since the correct identification of the factors is more relevant than having better-defined clusters.

To summarise, our reward aims to correctly identify each trajectory as belonging to the correct type of environment (e.g. the set of trajectories belonging to high-mass objects) as well as make the two sets of trajectories as distinguishable as possible.

An example may clarify how the F1 value and the clustering score are obtained in our experiments. We set up two sets of environments. In the first set, all objects have a low value of the mass, while in the second set, they have a high value. Each of these two values is defined as a range so that each object has a different value of the mass within the range. We then perform a robot movement action (performed in several timesteps), which will result in a displacement of the object. The trajectories are used to define a clustering function that identifies the two sets of objects. Finally, we perform the same action on new objects and use the obtained trajectories and the clustering function to identify them as belonging to the set of low or high masses. If we evaluate the results of these, we obtain on the one hand the F1 classification value (related to the number of correct classifications) and on the other hand the clustering score (indicating the distinguishability of the trajectories). In the next section, we will explain the optimisation process of the robot's actions, in order to maximise these scores. As we will see from the results of the experiments, in most of them it is possible to correctly classify objects with a score value of 1.0. In relation to this score, it is important to emphasise that the only goal is to assign one of the two possible categories correctly, so it is not surprising that this is not a problem. In fact, the original paper on this proposal does not include this score, probably for this reason. However, as we will see later, one of the aims of this project is to test the limits of the methodology by subjecting it to increasingly complex scenarios. In the last experiments, we will see how this classification is no longer correct in different scenarios. On the other hand, in all the experiments we also offer the value of the clustering score, which is never perfect, and which will serve to evaluate and compare in more detail all the results.

The full reward will be used both to drive the training of our RL system and to evaluate the performance of the results. In this respect, we will add details about this choice. The first of the elements considered, the Silhouette Score, is a common choice in the evaluation of the distinguishability of sets \cite{bagirov2023finding,punhani2022binning,shutaywi2021silhouette}. Additionally, it is the choice proposed in the original formulation of the method used, so it is useful for us to present its value in our experiments to compare our results with what would have been obtained in the original proposal. The second element, introduced as a novelty with respect to the original proposal, is the use of a score to evaluate the classification aspect. This is a fundamental point of our proposal since as we will see in the experiments carried out, especially in the last ones, when we confront this methodology with highly complex scenarios, it fails to correctly classify all the environments. This did not occur in the initial proposal, due to the simplicity of the scenarios analysed, which is similar to our first experiments, but it is an important element in this analysis of the limits and challenges of the methodology. Multiple standard metrics can be considered as classification scores, such as precision, recall, accuracy or F1. Our choice of F1 as the metric used represents an appropriate compromise, as it is the harmonic mean between precision and recall, which represents a good balance between them.

\subsection{Optimisation of interactions}
The last main element of the methodology is the definition of the approaches used to perform the search for optimal actions maximising the above reward. Here there is also a relevant difference in our methodology compared with the initial Causal Curiosity proposal. In the initial proposal, the Cross-Entropy Method (CEM) optimised Model Predictive Control Planner \cite{de2005tutorial,camacho2013model} is used. It works as follows: In each iteration, the planner proposes a set of plans sampled from a uniform distribution of the robot movements. Each plan is a sequence of movements, generated by applying control signals to the actuators in the joints of the robot. Using a horizon equal to 6 implies that during the experiment each joint receives 6 control signals, evenly spaced in time. A set of plans is proposed and each of the plans is executed in each of the environments. For each plan, we obtain a set of observations, determined by the positions of the object we interact with. The observations of each plan in the set of environments are used to obtain the reward for that plan. The subset of plans with the highest reward is selected to update the initial distribution from which we sampled the plans. This distribution is adjusted to fit the distribution related to those most successful plans. Once this is done, the process is repeated for another iteration. In this new iteration, the previously successful plans are more likely to be sampled. This will lead to the convergence of the plans that maximise the reward. As can be seen in this methodology, this is an open-loop control system, the actions are generated at the beginning of each experiment and thus are decoupled from the observations happening during the interaction.

This method is valid for simple scenarios but may be insufficient in more complex scenarios due to the lack of connection between action planning and observations. Therefore, in addition to this method, in this work we apply a PPO planner. This new planer has two fundamental differences with respect to the previous one. Firstly, we use a neural network to generate the actions of the robot. Secondly, this policy network is modified using a PPO method \cite{schulman2017proximal}. The process is carried out in the following steps: The policy network generates actions for the robot that similarly to the previous case are executed in the set of environments. As in the previous case, actions are generated as control signals applied regularly to the actuators in the joints of the robot. This produces a change of state in the environments with their corresponding observations, which together with their reward are used as input to produce an update in the policy network. Once this is done, a new set of actions is generated and the cycle is repeated. The basis of the implementation used has been provided by Stable Baselines\footnote{\url{https://stable-baselines3.readthedocs.io}}, on which we have made a number of modifications to implement the previous process. The execution of PPO is performed using multiprocessing vectorised environments. Each of the environments running in parallel corresponds to an environment $\textbf{e}^i$. Using wrappers we modified the actions so that all the environments reproduce the same actions (which by default does not happen since in general, the objective of parallelisation is to probe different actions of the same policy), and we introduced a shared value of the reward, calculated in the last timestep using the observations of all the environments (by default each environment only uses its own observations to calculate the reward, which in this case is insufficient since the reward is defined by comparison between environments). These modifications allow us to use a more efficient method such as PPO, while maintaining a similar approach to the original method.
In the original article, PPO is only used to carry out tasks subsequent to the estimation of the causal factors, while in our case this method is used directly for the estimation of the factors themselves. Throughout this paper, we will present in parallel the results using CEM and the results using PPO. The former corresponds to the methodology used in the original article, and the latter in this case. In this way, we will be able to compare the differences between both methodologies for each experiment.

\subsection{Implementation and experimental details}
Regarding the implementation details, the CEM planer has been implemented from the code of the Causal Curiosity repository\footnote{\label{causalcuriofootnote}\url{https://github.com/sumedh7/CausalCuriosity}}. The code has been optimised to allow multiprocessing and to allow the simultaneous variation of several causal factors. The output has also been modified to obtain the new reward proposed by us in this section. 

The experiments are carried out as follows. For each experiment, 20 environments are created, half of which corresponds to one of the ranges of the parameter to be estimated and half to the other range (e.g. light masses and heavy masses). An action is then executed in the environment in 6 movements. This action is executed in the 20 environments, obtaining 20 trajectories of the objects. The trajectories are used to define a clustering function, which allows to obtain the scores measuring the distinguishability of the parameter sets.

In each experiment, there are 100 replications of this process. As will be seen from the results presented in the next section, in several experiments the value of the scores is low. In the case of using this procedure in a real use case, the process could be optimised to further improve the score obtained. In our work, the aim is to show the limitations of the methodology in a comparative way, so we carried out all the experiments at the same optimisation levels, to allow the differences between the experiments to be shown, in order to understand the limits of the experiments.

The experiments can be reproduced by following these steps using the code provided in the original paper\footref{causalcuriofootnote}. It does not provide the value of the classification score, but this can be obtained directly from the trajectories by comparing them with the clustering function.

The hyperparameters relevant to this methodology are the following: environments = 20, plans per iteration = 5, iterations = 100, ratio of plans selected between updates = 0.4, frames per episode = 198, plan horizon = 6.
The PPO planer has been implemented from the above code and the PPO2 implementation of Stable Baselines, with the modifications mentioned above. The relevant hyperparameters are: environments = 20, discount factor gamma = 0.9995, entropy coefficient ent\_coef = 0, learning rate = 0.00005, value function coefficient vf\_coef = 0.5, maximum value for the gradient clipping max\_grad\_norm = 10, number of epochs when optimising the surrogate noptepochs = 4, iterations = 500, MlpPolicy network with 2 hidden layers with sizes 256 and 128, frames per episode = 198, plan horizon = 6.

These hyperparameters have been defined without the need for hypertuning processes because as mentioned above, the aim of the experiments is not to obtain the best possible result tuned to each scenario but to provide a comparative result between the different scenarios, in order to show the scenarios that offer the greatest challenges to the methodology. However, in real use cases, it may be relevant to set the parameters for better performance. For each dynamic system, a different set of parameters may be the most relevant in their effect on performance. Multiple studies analyse the effect of these parameters and propose ideas for optimal selection \cite{henderson2018deep,eimer2023hyperparameters,kiran2022hyperparameter}.
We recommend the use of libraries to optimise the search for suitable parameters. Optuna\footnote{\url{https://optuna.org/}}, Hyperopt\footnote{\url{https://hyperopt.github.io/hyperopt/}}, and Ray-Tune\footnote{\url{https://docs.ray.io/en/latest/tune/index.html}} are commonly used libraries for this purpose. They allow to parallelise searches, which significantly increases their efficiency. Another advantage they offer is the possibility of using more advanced search techniques such as Bayesian optimisation, in which the degree of uncertainty over the parameter space is quantified and used to select new regions to explore. This methodology is much more effective than a blind or a grid search.

The rationale for the selection of parameters to be estimated has been to offer the largest possible set of parameters that define the dynamics of the implemented case. The platform used for the simulations restricts the definition of solid objects\footnote{\url{https://causal-world.readthedocs.io/en/latest/modules/envs/envs.html#rigidobject}} to the parameters of mass, size, lateral friction and spinning friction, and the gravity parameter affecting the dynamics of the interaction between the robot and the object. As we will see in the results, especially in the more complex experiments, each type of parameter affects differently the dynamics of the experiments and thus the effectiveness of the methodology.

All the experiments will follow this methodology. In the next section, we describe the details of each of the experiments.

\section{Experiments}\label{sec:exp}

The experiments carried out correspond to each of the five research questions presented in Section \ref{sec:intro}.

In the following, we present each experiment and its results.

\subsection{Comparison of estimation of different causal factors}\label{subsec:fact_comparison}

The first experiment aims to provide a basis for comparison of subsequent more complex scenarios. Here we analyse the efficiency of the methods in correctly identifying causal factors using a wide range of values, and examine whether there are notable differences according to the type of causal factor at this level of experimentation.

\begin{figure}[ht]
    \centering
    \includegraphics[width=1.00\linewidth]{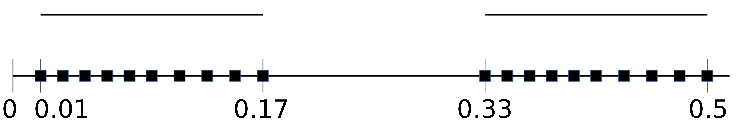}
    \caption{Mass values}
    \label{fig:exp1}
\end{figure}

In Figure \ref{fig:exp1} we can see a schematic representation of the values of one of these experiments, with the mass values organised into two groups represented by the two horizontal lines above. As can be seen in the figure, in this case, there is a set of objects with masses between 0.01 and 0.17, and another set of objects with masses between 0.33 and 0.5. We expect to find the sequence of interactions with the objects, which given a new object of unknown mass, maximises the correct identification of its mass (as belonging to one or the other range of masses).

In Table \ref{tab:acc_fact} we show the results of the experiment. We present the two scores independently: F1 classification and clustering score, and recall that the former indicates that the identification of the factors is properly performed, while the latter represents how distinguishable the clusters are from each other. As can be seen, this experiment does not present any identification problem, since all the F1 scores are 1.0, which means that all the trajectories are properly identified as belonging to one set of masses or the other. The results show similar clustering scores regardless of the factor under investigation. This shows that at this level the robustness in the estimation of all of them is similar, none of them seems to be more challenging in this initial experiment. As a comparison between the two optimisation strategies, if we take the mean value of the clustering scores, we obtain a value of $0.896\pm0.038$ for the CEM planner and $0.873\pm0.045$ for the PPO planner. However, we should remember that they represent qualitatively different parameters. This experiment provides some initial numbers to evaluate the accuracy of the estimation of different causal factors (RQ1) in a simple scenario and will be used for comparison in the following experiments when analysing more complex scenarios.

\begin{table*}[htp]
    \centering
    \begin{tabular}{ccccccc}

    \toprule
    \multirow{2}{*}{\textbf{Causal factor}} &   & \multirow{2}{*}{\textbf{Range}} & \multicolumn{2}{c}{\textbf{CEM planner}} & \multicolumn{2}{c}{\textbf{PPO planner}}\\
   \cmidrule(lr){4-5} \cmidrule(lr){6-7} 
     & & & \textbf{F1} & \textbf{Clust.} & \textbf{F1} & \textbf{Clust.}  \\
    \midrule
    \textbf{Mass} & F & [0.01,0.1733][0.3367,0.5] & 1.00 & 0.909 & 1.00 & 0.836  \\  
    \midrule
    \textbf{Size} & F & [0.05,0.0667][0.0833,0.1] & 1.00 & 0.879 & 1.00 & 0.815  \\    
    \midrule
    \textbf{Lateral Friction} & F & [0.1,0.4][0.7,1.0] & 1.00 & 0.949 & 1.00 & 0.889  \\    
    \midrule
    \textbf{Spinning Friction} & F & [0.001,0.334][0.667,1.0] & 1.00 & 0.846 & 1.00 & 0.914  \\  
    \midrule
    \textbf{Gravity} & F & [-1.0,-4.5][-8.0,-11.5] & 1.00 & 0.896 & 1.00 & 0.912  \\  
    \bottomrule
    \end{tabular}
            \caption{Comparison of estimation of different causal factors}
    \label{tab:acc_fact}
\end{table*}

\subsection{Granularity precision in causal factor determination}\label{subsec:granularity}

\begin{figure}[ht]
    \centering
    \includegraphics[width=1.00\linewidth]{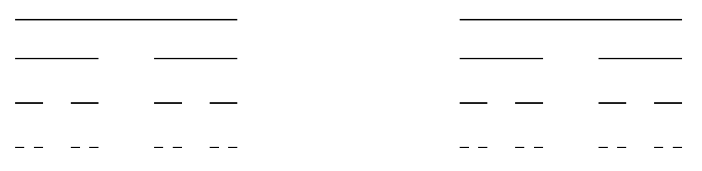}
    \caption{Successive bipartition of causal factor ranges. The last three levels are omitted}
    \label{fig:exp2}
\end{figure}

In this experiment, we analyse a more complex scenario, where our aim is to assign a more precise value to the causal factor and not just categorise it into two ranges. The objective is to understand how being more precise in the determination of the values (i.e. reducing the granularity of the considered ranges) affects the results (RQ2). Following the initial approach, what we do is take the two initial clusters and further split them into two other clusters. This procedure is repeated up to six times. In Figure \ref{fig:exp2} we see the representation of this procedure, and how the last clusters present a high precision in the knowledge of the value compared to the first ones (in the figure we represent only the first three bipartitions, instead of the six that we have performed). We note that although we follow a simple bipartition model here, the clusters can be organised with a uniform spacing at the desired level of precision, to better represent the parameter range. To clarify the level of precision we are talking about, consider that the ratio between the length of the range of values explored in the first experiment and this one is 0.0014. That is, if the strategy is successful, we are able to estimate each parameter with an accuracy three orders of magnitude higher.

Table \ref{tab:acc_gran} shows the results of this experiment. To synthesise the results, we present here values at the last level, and thus with the highest precision in identifying the factors. At this level, and according to the previous figure, there are 64 pairs of clusters. We present 2 pairs of clusters in the table, corresponding to the left beginning and right end of the range, and identified in the table respectively with the initials L and R. In this way, we can also take into account the effect of the factor being at one end or the other of the range (i.e. discerning between two light masses may be different to discerning between two heavy masses). To compare the change in the effectiveness of the experiment, we also include in the table the results of the first experiment over the full range, identified as F.

\begin{table*}[htp]
    \centering
    \begin{tabular}{ccccccc}
    \toprule
    \multirow{2}{*}{\textbf{Causal factor}} &   & \multirow{2}{*}{\textbf{Range}} & \multicolumn{2}{c}{\textbf{CEM planner}} & \multicolumn{2}{c}{\textbf{PPO planner}}\\
   \cmidrule(lr){4-5} \cmidrule(lr){6-7} 
     & & & \textbf{F1} & \textbf{Clust.} & \textbf{F1} & \textbf{Clust.}  \\
    \midrule
    \multirow{3}{*}{\textbf{Mass}} &  F & [0.01,0.1733][0.3367,0.5] &  1.00 &  0.909 &  1.00&  0.836 \\
      & L &  [0.01,0.0102][0.0104,0.0107] & 1.00 & 0.910 & 1.00 & 0.964   \\
      & R &  [0.4993,0.4996][0.4998,0.5] & 1.00 & 0.989 & 1.00 & 0.997   \\
    \midrule
    \multirow{3}{*}{\textbf{Size}} &  F & [0.05,0.0667][0.0833,0.1] &  1.00 &  0.879 &  1.00 &  0.815 \\
      & L & [0.05,0.05002][0.05005,0.05007] & 1.00 & 0.987 & 1.00 & 0.992   \\
      & R &  [0.09993,0.09995][0.09998,0.1] & 1.00 & 0.996 & 1.00 & 0.991   \\
    \midrule
    \multirow{3}{*}{\textbf{Lateral Friction}}  &  F  &  [0.1,0.4][0.7,1.0]  &  1.00  & 0.949  &  1.00  &  0.889  \\
      & L &  [0.1,0.1004][0.1008,0.1012] & 1.00 & 0.993 & 1.00 & 0.994   \\
      & R &  [0.9988,0.9992][0.9996,1.0] & 1.00 & 0.998 & 1.00 & 0.997   \\
    \midrule
    \multirow{3}{*}{\textbf{Spinning Friction}} &  F  & [0.001,0.334][0.667,1.0]& 1.00 & 0.846 & 1.00 & 0.914  \\
      & L &  [0.001,0.0015][0.0019,0.0024] & 1.00 & 0.955 & 1.00 & 0.993   \\
      & R &  [0.9986,0.9991][0.9995,1.0] & 1.00 & 0.803 & 1.00 & 0.999   \\
    \midrule
    \multirow{3}{*}{\textbf{Gravity}} &  F & [-1.0,-4.5][-8.0,-11.5]  &  1.00 &  0.896 & 1.00 &  0.912   \\
      & L &  [-1.0,-1.0048][-1.0096,-1.0144] & 1.00 & 0.879 & 1.00 & 0.933   \\
      & R &  [-11.4856,-11.4904][-11.4952,-11.5] & 1.00 & 0.998 & 1.00 & 0.999   \\
    \bottomrule
    \end{tabular}
    \caption{Analysis of granularity in the estimation of causal factors}
    \label{tab:acc_gran}
\end{table*}

From the results, we can see that even at this high level of precision it is possible to correctly estimate all causal factors. Additionally, we see in the value of the clustering score that in general the clusters are more defined than in the first experiment. It is important to note that this is not an effect derived from the score itself, since what the score measures, the relationship between the size of each cluster and the spacing between them, remains constant at all levels. What this result indicates is that the robot is able to find more efficient strategies when dealing with quasi-point distributions, than in the case of distributions with larger spreads, which is a positive outcome since this may be often the use case of interest (it is in general more useful to estimate parameters with high precision, and not just broad ranges). As a general evaluation, if we compare the clustering scores of the first and last level, they increase on average by 5.6\% and 6.7\% for the L and R cases respectively using the CEM planner. For the PPO planner case, the improvement is even much greater, with 12.0\% and 14.4\% for L and R respectively. 
We observe now a difference between the two methods used, CEM and PPO planners. In the previous experiment, the results of both methods were essentially equivalent. Now, when confronted with a more challenging experiment, we see that PPO shows better results in the clustering of the factors.

To better clarify the level of range accuracy achieved in this experiment, we can take the average value of each range as representative of the same: for example in the case of Spinning Friction, at the first level (F) we are discerning between a value of 0.17 and 0.83, while at the last level (L) we are able to distinguish between a friction value of 0.0012 and a value of 0.0021.

\subsection{Gap size effect in the determination of different factors}\label{subsec:gap}

\begin{figure}[ht]
    \centering
    \includegraphics[width=1.00\linewidth]{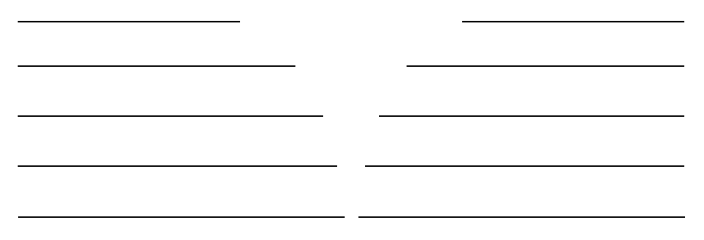}
    \caption{Reduction of gap size}
    \label{fig:exp3}
\end{figure}

In this case, the objective is to investigate how well-defined the ranges can be. Here we examine the effect in the results of reducing the gap between clusters in relation to their size (RQ3). Following the idea of the previous experiments, we start from an initial gap with the same size as each cluster and in each new experiment we divide this gap by half. We repeat this up to four times, going from a gap to cluster size ratio of 100\% to 4.2\% in the last case. Figure \ref{fig:exp3} presents these scenarios.

The results of the experiment are presented in Table \ref{tab:gap}. We include the case denoted as G with the shortest gap after 4 partitions and again the initial reference scenario denoted as F.

\begin{table*}[htp]
    \centering
    \begin{tabular}{ccccccc}
    \toprule
    \multirow{2}{*}{\textbf{Causal factor}} &   & \multirow{2}{*}{\textbf{Range}} & \multicolumn{2}{c}{\textbf{CEM planner}} & \multicolumn{2}{c}{\textbf{PPO planner}}\\
   \cmidrule(lr){4-5} \cmidrule(lr){6-7} 
     & & & \textbf{F1} & \textbf{Clust.} & \textbf{F1} & \textbf{Clust.}  \\
    \midrule
    \multirow{2}{*}{\textbf{Mass}} & F & [0.01,0.1733][0.3367,0.5] & 1.00 & 0.909 & 1.00 & 0.836  \\
      & G &  [0.01,0.2499][0.2601,0.5] & 1.00 & 0.774 & 0.90 & 0.627  \\
    \midrule
    \multirow{2}{*}{\textbf{Size}} & F & [0.05,0.0667][0.0833,0.1] & 1.00 & 0.879 & 1.00 & 0.815  \\
      & G &  [0.05,0.0745][0.0755,0.1] & 1.00 & 0.616 & 1.00 & 0.670  \\  
    \midrule
    \multirow{2}{*}{\textbf{Lateral Friction}} & F & [0.1,0.4][0.7,1.0] & 1.00 & 0.949 & 1.00 & 0.889  \\
      & G &  [0.1,0.5406][0.5594,1.0] & 1.00 & 0.630 & 1.00 & 0.672  \\
    \midrule
    \multirow{2}{*}{\textbf{Spinning Friction}} & F & [0.001,0.334][0.667,1.0] & 1.00 & 0.846 & 1.00 & 0.914  \\
      & G &  [0.001,0.4901][0.5109,1.0] & 0.95 & 0.600 & 1.00 & 0.616 \\
    \midrule
    \multirow{2}{*}{\textbf{Gravity}} & F & [-1.0,-4.5][-8.0,-11.5] & 1.00 & 0.896 & 1.00 & 0.912  \\
      & G &  [-1.0,-6.1406][-6.3594,-11.5] & 1.00 & 0.767 & 1.00 & 0.669  \\
    \bottomrule
    \end{tabular}
    \caption{Analysis of gap size effect in the estimation of causal factors}
    \label{tab:gap}
\end{table*}

As can be seen from the results, in this case, we encounter some difficulties. In the case of two of the causal factors (Spinning Friction and Mass), the identification of the factors is not always correct, obtaining a slightly lower F1 score, and in general, we observe in all of the factors a significant drop in the clustering score values, decreasing on average very similarly for the CEM and the PPO planners by 24.4\% and 25.3\% respectively. This is consistent with the results of the previous experiment in which we observed that the lower dispersion of the clusters improved the accuracy of the factor determination. It is important to note that although a lower clustering score may indicate a lower robustness of the results, in this experiment we are still able to identify correctly the causal factors in almost all cases. Therefore, although we begin to observe the limitations of this methodology, it is still satisfactory for the design of an effective exploration strategy.

\subsection{Multiple causal factors estimation}\label{subsec:multiple}

\begin{figure}[ht]
    \centering
    \includegraphics[width=1.00\linewidth]{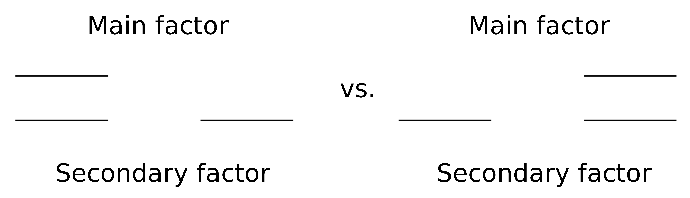}
    \caption{Variation of two causal factors}
    \label{fig:exp4}
\end{figure}

In this section, we investigate the effect of simultaneously modifying several causal factors (RQ4). So far we have considered ideal experimental scenarios, in which we can make interventions on one variable while holding all other variables constant. Here we investigate less gracious real-world scenarios where this is not possible. In this case, two causal factors vary. Our objective is to determine the value of one of them (the main causal factor), which continues to vary between two ranges of values as in the previous experiments. But now, in each of these two groups of scenarios, there is additionally a variation of the other factor (the secondary causal factor) taking place as well between two ranges of values. E.g. the two groups to be clustered and identified as different could be `high mass with large and small sizes' versus `low mass with large and small sizes'. In Figure \ref{fig:exp4} we represent this experiment. To correctly understand this scenario we highlight that the two factors do not vary in a correlated way (in the previous example, it is not `high mass and large size' versus `low mass and small size'). Instead, the secondary factor can be seen as a noise effect that makes it difficult to determine the first one. The variation of the second causal factor is randomised between its two possible ranges, but maintains the same order with respect to each main causal factor and also with respect to all the exploration strategies in each experiment.

We report the results in Tables \ref{tab:multi_cem_f1}, \ref{tab:multi_cem_cluster}, \ref{tab:multi_ppo_f1} and \ref{tab:multi_ppo_cluster}. In each table, the first column represents the main causal factor, and the first row the secondary causal factor. We have included all combinations, except for Mass and Gravity whose effects partially counteract each other. For clarity, we present F1 and clustering scores separately. In this case, we do not observe a general trend for all the factors that allow us to quantify the overall improvement or worsening, but rather we obtain qualitative results depending on the combination of factors to be studied.

\begin{table*}[htp]
    
    \centering
    \begin{tabular}{c|ccccc}
    \toprule
    & \textbf{Mass} & \textbf{Size} & \textbf{Lat. Frict.} & \textbf{Spin. Frict.} & \textbf{Gravity} \\
    \midrule
    \textbf{Mass} &  & 0.85 & 0.95 & 1.00 &  \\
    \textbf{Size}  & 1.00 &  & 1.00 & 1.00 & 1.00 \\
    \textbf{Lat. Frict.}  & 0.85 & 1.00 &  & 1.00 & 1.00 \\
    \textbf{Spin. Frict.}  & 0.65 & 0.55 & 0.75 &  & 0.65 \\
    \textbf{Gravity}  &  & 0.90 & 1.00 & 1.00 &  \\
    \bottomrule
    \end{tabular}
    
\caption{Analysis of multiple causal factors estimation. CEM planner. F1 score.}
    \label{tab:multi_cem_f1}
\end{table*}

\begin{table*}[htp]
    \centering
    \begin{tabular}{c|ccccc}
    \toprule
    & \textbf{Mass} & \textbf{Size} & \textbf{Lat. Frict.} & \textbf{Spin. Frict.} & \textbf{Gravity} \\
    \midrule
    \textbf{Mass}  &  & 0.766 & 0.722 & 0.883 &  \\
    \textbf{Size}  & 0.804 &  & 0.848 & 0.827 & 0.788 \\
    \textbf{Lat. Frict.}  & 0.474 & 0.641 &  & 0.832 & 0.877 \\
    \textbf{Spin. Frict.}  & 0.482 & 0.701 & 0.544 &  & 0.725 \\
    \textbf{Gravity}  &  & 0.530 & 0.731 & 0.839 &  \\
    \bottomrule
    \end{tabular}
    \caption{Analysis of multiple causal factors estimation. CEM planner. Clustering score.}
    \label{tab:multi_cem_cluster}
\end{table*}

\begin{table*}[htp]
    \centering
    \begin{tabular}{c|ccccc}
    \toprule
    & \textbf{Mass} & \textbf{Size} & \textbf{Lat. Frict.} & \textbf{Spin. Frict.} & \textbf{Gravity} \\
    \midrule
    \textbf{Mass}  &  & 0.95 & 1.00 & 1.00 &  \\
    \textbf{Size}  & 1.00 &  & 1.00 & 1.00 & 1.00 \\
    \textbf{Lat. Frict.}  & 0.80 & 0.60 &  & 1.00 & 1.00 \\
    \textbf{Spin. Frict.}  & 0.65 & 0.55 & 1.00 &  & 0.75 \\
    \textbf{Gravity}  &  & 0.75 & 1.00 & 1.00 &  \\
    \bottomrule
    \end{tabular}
    \caption{Analysis of multiple causal factors estimation. PPO planner. F1 score.}
    \label{tab:multi_ppo_f1}
\end{table*}

\begin{table*}[htp]
    \centering 
    \begin{tabular}{c|ccccc}
    \toprule
     & \textbf{Mass} & \textbf{Size} & \textbf{Lat. Frict.} & \textbf{Spin. Frict.} & \textbf{Gravity} \\
     \midrule
    \textbf{Mass} &  & 0.608 & 0.615 & 0.807 &  \\
    \textbf{Size} & 0.797 &  & 0.800 & 0.811 & 0.743 \\
    \textbf{Lat. Frict.} & 0.362 & 0.595 &  & 0.873 & 0.590 \\
    \textbf{Spin. Frict.} & 0.696 & 0.767 & 0.693 &  & 0.404 \\
    \textbf{Gravity}  &  & 0.681 & 0.832 & 0.885 &  \\
    \bottomrule
    \end{tabular}
    \caption{Analysis of multiple causal factors estimation. PPO planner. Clustering score.}
    \label{tab:multi_ppo_cluster}
\end{table*}

It can be seen from the F1 score results that in several of these experiments, it is not possible to correctly identify the main causal factor. We are confronted with the first group of experiments that clearly reflect the limitations of the methodology used.

We also observe that the ability to identify factors changes significantly depending on the chosen factor, something that has also not been observed so far in the previous experiments. In particular, we see for example how it is very difficult to determine the Spinning Friction value, while it is easy to identify the Size independently of the noise in the other factors. The latter makes sense, as dynamics that distinguish between different sizes can be determined mainly by the point where the force is applied, which is independent of the other factors, while in other cases, the interaction of the two factors may be correlated (at least depending on the specific type of movement triggered by each exploration strategy). We also note that Lateral and Spinning Friction are both less relevant as secondary factors.

It is therefore relevant when studying the dynamics of a particular system to take into account the specific hierarchy that can be established between the different factors, in order to be able to develop suitable experimentation strategies for their identification.

In this case, it should be noted that the clustering score should be considered in a secondary way with respect to the F1 score. That is, a high clustering score indicates clusters with little spread, but it is still of little relevance if part of the elements of each cluster are erroneous and belong to the opposite cluster.
Nevertheless, this score is useful to see some of the previous cases in more detail. For example, Mass can be correctly identified as the main factor with respect to Lateral or Spinning Friction. But in the latter case the clustering value is much higher than in the former. This is consistent with the fact that Spinning Friction obtains the lowest values of identification as a main factor.

\subsection{Causally related and confounding causal factors determination}\label{subsec:confounding}

\begin{figure}[ht]
    \centering
    \includegraphics[width=0.70\linewidth]{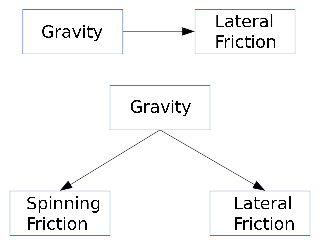}
    \caption{Causal Directed Acyclic Graphs between the factors}
    \label{fig:dags}
\end{figure}

The last experiment represents the most complex scenario of this work. We allow several factors to vary during the experiment, as in the previous case, but now the factors are causally related to each other (RQ5), while in the last experiment the factors were uncorrelated. We analyse two causal models, in the first one, a factor (Gravity) is the cause of a second factor (Lateral Friction). In the second model, one factor (Gravity) is a confounding variable with respect to two other factors (Lateral and Spinning Friction). In Figure \ref{fig:dags} we represent the Causal DAG corresponding to these two cases, where we omit the parameters that do not vary and the observations. These types of DAG represent two key scenarios with respect to the causal study of systems with two and three variables.

To analyse these causal relationships, and given the experimental framework used, we propose to model these scenarios by means of discrete Additive Noise Models (ANM) \cite{peters2011causal}. The first causal model is defined by the following equations:

\begin{equation}
\begin{split}
\mathbf{L} = f_L\left(\mathbf{G}\right) + \mathbf{N_1} \mathrm{~and~} \mathbf{N_1}~\bot~\mathbf{G} \\
\mathbf{G} \neq f_G\left(\mathbf{L}\right) + \mathbf{N_2} \mathrm{~and~} \mathbf{N_2}~\bot~\mathbf{L} 
\end{split}
\end{equation}

where $\mathbf{L}$ and $\mathbf{G}$ represent the Gravity and Lateral Friction values, $f_L$ and $f_G$ are independent functions, and $\mathbf{N_n}$ are independent noise variables for each factor. 

As explicitly stated in this equation, in the case of these ANMs the above equality only holds in the causal direction.

\begin{figure}[ht]
    \centering
    \includegraphics[width=1.00\linewidth]{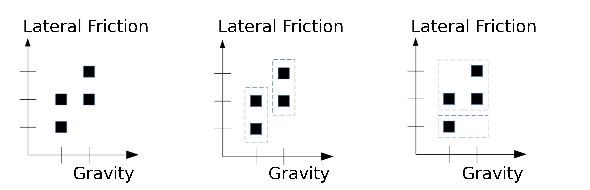}
    \caption{Values for the first causal model, and two clustering scenarios (C1 and C2)}
    \label{fig:exp5a}
\end{figure}

The values of the factors corresponding to the first model are shown in Figure \ref{fig:exp5a}. In the left part of the figure, we see the values, where each square indicates that each variable does not have a point distribution but represents a range of values (the line of previous representations is now, therefore, a square). Next, we observe two possible ways of clustering these values, identifying the value of Gravity in the second figure, and of Lateral Friction in the third.

The second causal model is defined by the following equations:

\begin{equation}
\begin{split}
\mathbf{L} = h_L\left(\mathbf{G}\right) + \mathbf{N_3} \mathrm{~and~} \mathbf{N_3}~\bot~\mathbf{G}  \\
\mathbf{S} = h_S\left(\mathbf{G}\right) + \mathbf{N_4} \mathrm{~and~} \mathbf{N_4}~\bot~\mathbf{G}  \\
\end{split}
\end{equation}

where $\mathbf{S}$ represents the Spinning Friction, and we omit the anticausal inequalities.

\begin{figure}[ht]
    \centering
    \includegraphics[width=1.00\linewidth]{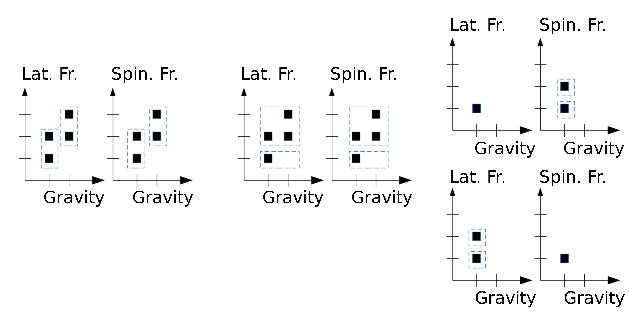}
    \caption{Four clustering scenarios for the second causal model (C3 to C6)}
    \label{fig:exp5b}
\end{figure}

This second experiment is represented in Figure \ref{fig:exp5b}. As we can see, this can be represented as two parallel relationships in which in both cases Gravity is the main cause of the other factors. In this figure, we represent four possible ways of clustering the values. The first two represent the main identification of the values: Gravity in the first case and the other two factors simultaneously in the second case as they are correlated. In the last two graphs (above and below) we represent the case in which once both factors (Lateral and Spinning Friction) are identified we apply a new strategy to differentiate between one and the other, in case their dispersion is not equal.

The results of these experiments are presented in Table \ref{tab:causal_conf}.

\begin{table*}[htp]
    \centering 
    \begin{tabular}{cccccc}
    \toprule
    \multirow{2}{*}{\textbf{Causal factors}} & \multirow{2}{*}{\textbf{Range}} & \multicolumn{2}{c}{\textbf{CEM planner}} & \multicolumn{2}{c}{\textbf{PPO planner}}\\
   \cmidrule(lr){3-4} \cmidrule(lr){5-6} 
     & & \textbf{F1} & \textbf{Clust.} & \textbf{F1} & \textbf{Clust.}  \\
    \midrule    
    \textbf{C1}~~ G$\rightarrow$L   & G:[-1.0,-4.5][-8.0,-11.5] & \multirow{2}{*}{1.00} & \multirow{2}{*}{0.848} & \multirow{2}{*}{1.00} & \multirow{2}{*}{0.903}  \\   
    cluster G & L:[0.1,0.28][0.46,0.64][0.82,1.0]  &  &  &  &   \\   
    \midrule    
    \textbf{C2}~~ G$\rightarrow$L    & G:[-1.0,-4.5][-8.0,-11.5] & \multirow{2}{*}{1.00} & \multirow{2}{*}{0.626} & \multirow{2}{*}{1.00} & \multirow{2}{*}{0.740}  \\   
    cluster L & L:[0.1,0.28][0.46,0.64][0.82,1.0]  &  &  &  &   \\   
    \midrule    
    \multirow{2}{*}{\textbf{C3}~~ G$\rightarrow$L,S }   & G:[-1.0,-4.5][-8.0,-11.5] & \multirow{4}{*}{1.00} & \multirow{4}{*}{0.811} & \multirow{4}{*}{1.00} & \multirow{4}{*}{0.803}  \\   
    & L:[0.1,0.28][0.46,0.64][0.82,1.0]  &  &  &  &   \\   
    \multirow{2}{*}{cluster G }   & S:[0.001,0.2008][0.4006,0.6004]   &  &  &  &   \\   
    & [0.8002,1.0]  &  &  &  &   \\   
    \midrule    
    \multirow{2}{*}{\textbf{C4}~~ G$\rightarrow$L,S }  & G:[-1.0,-4.5][-8.0,-11.5] & \multirow{4}{*}{1.00} & \multirow{4}{*}{0.621} & \multirow{4}{*}{1.00} & \multirow{4}{*}{0.698}  \\   
   & L:[0.1,0.28][0.46,0.64][0.82,1.0]  &  &  &  &   \\   
    \multirow{2}{*}{cluster L,S }    & S:[0.001,0.2008][0.4006,0.6004]   &  &  &  &   \\   
      & [0.8002,1.0]  &  &  &  &   \\   
    \midrule 
    \multirow{2}{*}{\textbf{C5}~~ G$\rightarrow$L,S }  & G:[-1.0,-4.5] & \multirow{3}{*}{1.00} & \multirow{3}{*}{0.691} & \multirow{3}{*}{1.00} & \multirow{3}{*}{0.846}  \\   
    &  L:[0.1,0.28][0.46,0.64]  &  &  &  &   \\   
    cluster L & S:[0.001,0.2008]  &  &  &  &   \\   
    \midrule    
    \multirow{2}{*}{\textbf{C6}~~ G$\rightarrow$L,S } & G:[-1.0,-4.5]& \multirow{3}{*}{0.95} & \multirow{3}{*}{0.927} & \multirow{3}{*}{1.00} & \multirow{3}{*}{0.800}  \\   
   & L:[0.1,0.28]  &  &  &  &   \\   
     cluster S  & S:[0.001,0.2008][0.4006,0.6004]  &  &  &  &   \\       
    \bottomrule
    \end{tabular}
    \caption{Analysis of causally related and confounding causal factors estimation}
    \label{tab:causal_conf}
\end{table*}

It can be seen from the results that in most cases it is possible to correctly identify the values of the causal factors. In relation to the first causal model with two variables (results C1 and C2 in the table), we obtain a slightly lower value for the second result, compatible with the results of the previous experiment and the fact that Lateral Friction was clustered in that case. In relation to the second model, we obtain similar results both in the first factor identification (C3 and C4) and in the finer discrimination between Lateral and Spinning Friction (C5 and C6). We obtain again slightly better results using the PPO planner.

It is important to clarify that these parameter identification strategies only allow us to determine the values of the factors. To determine the causal relationships between the variables we must apply causal structure identification techniques \cite{vowels2022d,glymour2019review,hasan2023survey} to the results. In this case, expert knowledge about the system could allow us to postulate an ANM, and values following a relationship similar to Figures \ref{fig:exp5a} or \ref{fig:exp5b} would indicate what the causal relationship is. However, we have presented a simple model in this work; in a real-world scenario, we would need a much larger collection of values to be able to determine this causal direction with certainty. Nevertheless, this simple model gives us a hopeful intuition about the scope of the Causal Curiosity method and the possibility of tackling the problem of causal structure identification.

\section{Discussion}\label{sec:discussion}

From the results of the experiments carried out, we can begin to glimpse the potential and limitations of Causal Curiosity. In simple scenarios where it is possible to vary a single parameter independently of the others, and where the ranges of variation are well defined (sections \ref{subsec:fact_comparison} and \ref{subsec:granularity}), this methodology works very efficiently and robustly in estimating the parameters that define our system. Under these conditions, it is also possible to estimate the parameters with very high precision (Section \ref{subsec:granularity}). The downside is that, as in many RL experiments, the method used here may involve a high number of experiments (directly proportional to the desired level of precision). In both scenarios, it is possible to apply a simple search methodology for exploration strategies. However, in order to reduce the number of experiments, an interesting line of future work opens up in which to design more optimal partitioning strategies than the bi-partitioning carried out here. In this experiment, we observe how we can determine the parameters in a range three orders of magnitude finer and at the same time obtain an improvement in the clustering score on average up to 14.4\% for the more efficient PPO planer.
One specific possibility to work on for the future line of work mentioned could be to use the clustering score as a measure of uncertainty of the parameter estimation in a Bayesian-like approach. In the methodology explored here, the range of parameter exploration is divided at each iteration in an agnostic manner, only driven by the segment offering the highest score. In this other proposed line of work, the score may serve as a weight that quantifies the certainty with which we know the parameter. This implies that sometimes it may be more relevant to explore unknown areas to reduce our lack of knowledge rather than following a greedy approach.

On the other hand, we observe the limits of the methodology in two situations. The first occurs when the ranges of the parameters to be identified are close to each other (section \ref{subsec:gap}). Here, by reducing the gap between them to about 4\% of their size, we observe a reduction in the clustering score of about 25\%. This is not an unsolvable constraint, but it reveals limits to be taken into account. In this case, we can divide these wide ranges with a small gap in between into multiple well-defined sub-ranges with a wide gap (similar to what was done in Section \ref{subsec:granularity}), although this obviously increases the number of explorations to be carried out. 

The second challenging situation represents a scenario in which ideal experimental conditions are not possible, and at least a second factor varies at the same time as the first (section \ref{subsec:multiple}). Here we see how this case represents a hard limit of our methodology, and even with the simple experimental scenario employed here we observe cases in which it is not possible to correctly estimate the parameters. We observe the important effect of choosing to estimate one parameter or another. This second situation is therefore totally dependent on the type of dynamic system to be investigated. It also points to a second interesting line of future research, in which to compare very different systems in order to develop robust Causal Curiosity methodologies both in relation to the systems investigated and to the choice of causal factors. At the end of this section, we propose a number of use cases where this methodology could be applied. The diversity of domains suggested is a good example of dynamical systems governed by very different equations, which would allow a more comprehensive understanding of the effect of the combination of parameters in the methodology.

The last scenario we have analysed proposes a simple modelling of a scenario with factors with causal relationships between them (Section \ref{subsec:confounding}). This involves taking a step beyond the previous section by moving even further away from the ideal experimental conditions. We observe again a better result for the PPO planer. We should not extrapolate the conclusions of this scenario too far, again due to the dependence on system, factors, and in this case causal relationships. Nevertheless, the positive outcomes of this first experiment open a door to the possibility of being successful in other situations. Additionally, it serves as a starting point for the development of experiments where the impact of these causal relationships between factors is analysed more comprehensively. This opens a path as a third future line of research in Causal ML combining the field of Causal structure identification \cite{vowels2022d,glymour2019review,hasan2023survey} with the type of Reinforcement Learning experimentation carried out in this work. The specific proposal for future work would be a methodology in which to sequentially combine parameter estimation experiments with structure identification experiments. One of the tools available in the latter domain is the study of interventions as a tool to clarify the causal relationship between parameters. In simple terms, by observing correlations between variables we can design specific experiments to differentiate correlations from true causal relationships. In this case, the value of the parameters is irrelevant, the interest is only in the relationships (the arrows in the causal diagrams). By alternating these experiments with the ones proposed in this paper, we can in a combined way increase our knowledge of both elements, the value of the parameters and their relationships, and reduce their shared interplay.

Taking these results into account, we can consider how they affect the generalisability of this methodology to other scenarios than the ones explored in this work. First, we see that it is directly generalisable to any other dynamical system in which we only need to estimate one parameter. The methodology allows us to identify the interaction in the system that maximises the accuracy in this estimation, and to increase the precision of the value obtained, reducing the estimation range as much as required. The price to pay for increasing the accuracy is the increase in the number of interactions to be carried out. The number of interactions grows linearly with the number of times we want to subdivide the value range of a parameter. We can provide an example of a scenario where this methodology would be applicable. Let's assume the case of an autonomous vehicle driving in rough terrain. In this case, we can identify certain internal parameters of the system that are likely to be particularly affected, such as the wheel pressure, for example. The application of the methodology will produce the series of movements that will most accurately identify one pressure value or another. Independently, we can apply the methodology again to estimate another parameter, such as the orientation of the wheels relative to the vehicle axle, producing another set of movements.

The limitations observed in the last experiments show in turn the situations in which the application of the method can not be generalised. For instance, in the last example, let us consider the scenario in which we want to identify an environmental parameter external to the system such as the coefficient of friction of the ground, and also an internal parameter such as the brake wear. In this case, the two parameters are confounding variables (unlike the previous one in which the two parameters were mutually independent) so that for example when observing a prolonged displacement when braking we would not be able to identify whether the effect is produced by brake wear or a low coefficient of friction of the ground. Generalisability is therefore assured in the case of a single parameter, and in the case of multi-parameter estimation depends on the relationship between the parameters of each system. For this main limitation of the methodology, as a specific proposal for a fourth future line of work, we can consider a more advanced version of the methodology in which the reward includes a factor that precisely maximises the distinguishability between related parameters. In this case, experiments should include a minimum of two unknown parameters, but ideally more, and analyse the ability of the system to autonomously identify the combinations that are distinguishable from those that are not. As mentioned above, this opens up a line of work that connects directly to the field of causal structure identification as a subdomain of causal analysis, which would be a very fruitful addition to our line of work.

This methodology is especially useful in cases where the user cannot directly apply a measurement tool to estimate the relevant parameter. An exemplary case is that of autonomous systems. We have mentioned the case of an autonomous driving vehicle, other cases can be UAVs and their interactions with the environment, and robotic space exploration systems such as satellites or planetary rovers. On the other hand, the methodology is also relevant in the case of existing parameter measurement tools, as it offers a transversal procedure to corroborate the value of the measurement and thus ensure its validity. This is especially important in systems that require redundancy in the estimation of their states, such as aircrafts or other aerospace applications.

\section{Conclusions}\label{sec:conclusions}

In this project, we have analysed the limits and challenges presented by Causal Curiosity as a methodology to identify causal factors in a dynamic system, focusing on the case of a
robotic manipulator interacting with unknown objects. We have presented different experimental scenarios to explore the effectiveness and scope of the methodology in relation to: the analysis of different types of factors, the definition of their ranges in terms of precision and granularity, and the interaction of multiple factors in an uncorrelated way or through causal relationships. We have also presented comparative results between two search methods for factor estimation strategies, with typically better results from our proposed PPO planer. The results obtained show us both the limitations and the extensive potential of this methodology, and offer us concrete proposals for designing effective exploration strategies that can be applied to more complex frameworks or real-world use cases.

\section*{Declarations}

\begin{itemize}
\item Conflict of interest. The authors declare that there is neither funding nor conflict of interest.
\item Data Availability. All data included in this study are available upon request by contact with the corresponding author.
\end{itemize}

\bibliography{causalcuriosity}


\begin{thebibliography}{94}
\ifx \bisbn   \undefined \def \bisbn  #1{ISBN #1}\fi
\ifx \binits  \undefined \def \binits#1{#1}\fi
\ifx \bauthor  \undefined \def \bauthor#1{#1}\fi
\ifx \batitle  \undefined \def \batitle#1{#1}\fi
\ifx \bjtitle  \undefined \def \bjtitle#1{#1}\fi
\ifx \bvolume  \undefined \def \bvolume#1{\textbf{#1}}\fi
\ifx \byear  \undefined \def \byear#1{#1}\fi
\ifx \bissue  \undefined \def \bissue#1{#1}\fi
\ifx \bfpage  \undefined \def \bfpage#1{#1}\fi
\ifx \blpage  \undefined \def \blpage #1{#1}\fi
\ifx \burl  \undefined \def \burl#1{\textsf{#1}}\fi
\ifx \doiurl  \undefined \def \doiurl#1{\url{https://doi.org/#1}}\fi
\ifx \betal  \undefined \def \betal{\textit{et al.}}\fi
\ifx \binstitute  \undefined \def \binstitute#1{#1}\fi
\ifx \binstitutionaled  \undefined \def \binstitutionaled#1{#1}\fi
\ifx \bctitle  \undefined \def \bctitle#1{#1}\fi
\ifx \beditor  \undefined \def \beditor#1{#1}\fi
\ifx \bpublisher  \undefined \def \bpublisher#1{#1}\fi
\ifx \bbtitle  \undefined \def \bbtitle#1{#1}\fi
\ifx \bedition  \undefined \def \bedition#1{#1}\fi
\ifx \bseriesno  \undefined \def \bseriesno#1{#1}\fi
\ifx \blocation  \undefined \def \blocation#1{#1}\fi
\ifx \bsertitle  \undefined \def \bsertitle#1{#1}\fi
\ifx \bsnm \undefined \def \bsnm#1{#1}\fi
\ifx \bsuffix \undefined \def \bsuffix#1{#1}\fi
\ifx \bparticle \undefined \def \bparticle#1{#1}\fi
\ifx \barticle \undefined \def \barticle#1{#1}\fi
\bibcommenthead
\ifx \bconfdate \undefined \def \bconfdate #1{#1}\fi
\ifx \botherref \undefined \def \botherref #1{#1}\fi
\ifx \url \undefined \def \url#1{\textsf{#1}}\fi
\ifx \bchapter \undefined \def \bchapter#1{#1}\fi
\ifx \bbook \undefined \def \bbook#1{#1}\fi
\ifx \bcomment \undefined \def \bcomment#1{#1}\fi
\ifx \oauthor \undefined \def \oauthor#1{#1}\fi
\ifx \citeauthoryear \undefined \def \citeauthoryear#1{#1}\fi
\ifx \endbibitem  \undefined \def \endbibitem {}\fi
\ifx \bconflocation  \undefined \def \bconflocation#1{#1}\fi
\ifx \arxivurl  \undefined \def \arxivurl#1{\textsf{#1}}\fi
\csname PreBibitemsHook\endcsname

\bibitem[\protect\citeauthoryear{Sch\"{o}lkopf}{2022}]{scholkopf2022causality}
\begin{bbook}
\bauthor{\bsnm{Sch\"{o}lkopf}, \binits{B.}}:
\bbtitle{Causality for Machine Learning},
\bedition{1}st edn.,
pp. \bfpage{765}--\blpage{804}.
\bpublisher{Association for Computing Machinery},
\blocation{New York, NY, USA}
(\byear{2022}).
\doiurl{10.1145/3501714.3501755}
\end{bbook}
\endbibitem

\bibitem[\protect\citeauthoryear{Schölkopf et~al.}{2021}]{scholkopf2021toward}
\begin{barticle}
\bauthor{\bsnm{Schölkopf}, \binits{B.}},
\bauthor{\bsnm{Locatello}, \binits{F.}},
\bauthor{\bsnm{Bauer}, \binits{S.}},
\bauthor{\bsnm{Ke}, \binits{N.R.}},
\bauthor{\bsnm{Kalchbrenner}, \binits{N.}},
\bauthor{\bsnm{Goyal}, \binits{A.}},
\bauthor{\bsnm{Bengio}, \binits{Y.}}:
\batitle{Toward causal representation learning}.
\bjtitle{Proceedings of the IEEE}
\bvolume{109}(\bissue{5}),
\bfpage{612}--\blpage{634}
(\byear{2021})
\doiurl{10.1109/JPROC.2021.3058954}
\end{barticle}
\endbibitem

\bibitem[\protect\citeauthoryear{Locatello et~al.}{2019}]{locatello2019challenging}
\begin{bchapter}
\bauthor{\bsnm{Locatello}, \binits{F.}},
\bauthor{\bsnm{Bauer}, \binits{S.}},
\bauthor{\bsnm{Lucic}, \binits{M.}},
\bauthor{\bsnm{Raetsch}, \binits{G.}},
\bauthor{\bsnm{Gelly}, \binits{S.}},
\bauthor{\bsnm{Sch{\"o}lkopf}, \binits{B.}},
\bauthor{\bsnm{Bachem}, \binits{O.}}:
\bctitle{Challenging common assumptions in the unsupervised learning of disentangled representations}.
In: \bbtitle{International Conference on Machine Learning},
pp. \bfpage{4114}--\blpage{4124}
(\byear{2019}).
\bcomment{PMLR}
\end{bchapter}
\endbibitem

\bibitem[\protect\citeauthoryear{Suter et~al.}{2019}]{suter2019robustly}
\begin{bchapter}
\bauthor{\bsnm{Suter}, \binits{R.}},
\bauthor{\bsnm{Miladinovic}, \binits{D.}},
\bauthor{\bsnm{Sch{\"o}lkopf}, \binits{B.}},
\bauthor{\bsnm{Bauer}, \binits{S.}}:
\bctitle{Robustly disentangled causal mechanisms: Validating deep representations for interventional robustness}.
In: \bbtitle{International Conference on Machine Learning},
pp. \bfpage{6056}--\blpage{6065}
(\byear{2019}).
\bcomment{PMLR}
\end{bchapter}
\endbibitem

\bibitem[\protect\citeauthoryear{Sch\"{o}lkopf et~al.}{2012}]{scholkopf2012causal}
\begin{bchapter}
\bauthor{\bsnm{Sch\"{o}lkopf}, \binits{B.}},
\bauthor{\bsnm{Janzing}, \binits{D.}},
\bauthor{\bsnm{Peters}, \binits{J.}},
\bauthor{\bsnm{Sgouritsa}, \binits{E.}},
\bauthor{\bsnm{Zhang}, \binits{K.}},
\bauthor{\bsnm{Mooij}, \binits{J.}}:
\bctitle{On causal and anticausal learning}.
In: \bbtitle{Proceedings of the 29th International Coference on International Conference on Machine Learning}.
\bsertitle{ICML'12},
pp. \bfpage{459}--\blpage{466}.
\bpublisher{Omnipress},
\blocation{Madison, WI, USA}
(\byear{2012})
\end{bchapter}
\endbibitem

\bibitem[\protect\citeauthoryear{Kilbertus et~al.}{2018}]{kilbertus2018generalization}
\begin{botherref}
\oauthor{\bsnm{Kilbertus}, \binits{N.}},
\oauthor{\bsnm{Parascandolo}, \binits{G.}},
\oauthor{\bsnm{Sch{\"o}lkopf}, \binits{B.}}:
Generalization in anti-causal learning.
arXiv preprint arXiv:1812.00524
(2018)
\end{botherref}
\endbibitem

\bibitem[\protect\citeauthoryear{Lu et~al.}{2018}]{lu2018deconfounding}
\begin{botherref}
\oauthor{\bsnm{Lu}, \binits{C.}},
\oauthor{\bsnm{Sch{\"o}lkopf}, \binits{B.}},
\oauthor{\bsnm{Hern{\'a}ndez-Lobato}, \binits{J.M.}}:
Deconfounding reinforcement learning in observational settings.
arXiv preprint arXiv:1812.10576
(2018)
\end{botherref}
\endbibitem

\bibitem[\protect\citeauthoryear{Singla and Feizi}{2022}]{singla2021salient}
\begin{bchapter}
\bauthor{\bsnm{Singla}, \binits{S.}},
\bauthor{\bsnm{Feizi}, \binits{S.}}:
\bctitle{Salient imagenet: How to discover spurious features in deep learning?}
In: \bbtitle{International Conference on Learning Representations}
(\byear{2022})
\end{bchapter}
\endbibitem

\bibitem[\protect\citeauthoryear{Beery et~al.}{2018}]{beery2018recognition}
\begin{bchapter}
\bauthor{\bsnm{Beery}, \binits{S.}},
\bauthor{\bsnm{Van~Horn}, \binits{G.}},
\bauthor{\bsnm{Perona}, \binits{P.}}:
\bctitle{Recognition in terra incognita}.
In: \bbtitle{Proceedings of the European Conference on Computer Vision (ECCV)},
pp. \bfpage{456}--\blpage{473}
(\byear{2018})
\end{bchapter}
\endbibitem

\bibitem[\protect\citeauthoryear{Wang et~al.}{2022}]{wang2022causal}
\begin{bchapter}
\bauthor{\bsnm{Wang}, \binits{Z.}},
\bauthor{\bsnm{Xiao}, \binits{X.}},
\bauthor{\bsnm{Xu}, \binits{Z.}},
\bauthor{\bsnm{Zhu}, \binits{Y.}},
\bauthor{\bsnm{Stone}, \binits{P.}}:
\bctitle{Causal dynamics learning for task-independent state abstraction}.
In: \bbtitle{International Conference on Machine Learning},
pp. \bfpage{23151}--\blpage{23180}
(\byear{2022}).
\bcomment{PMLR}
\end{bchapter}
\endbibitem

\bibitem[\protect\citeauthoryear{De~Haan et~al.}{2019}]{de2019causal}
\begin{botherref}
\oauthor{\bsnm{De~Haan}, \binits{P.}},
\oauthor{\bsnm{Jayaraman}, \binits{D.}},
\oauthor{\bsnm{Levine}, \binits{S.}}:
Causal confusion in imitation learning.
Advances in neural information processing systems
\textbf{32}
(2019)
\end{botherref}
\endbibitem

\bibitem[\protect\citeauthoryear{Ortega et~al.}{2021}]{ortega2021shaking}
\begin{botherref}
\oauthor{\bsnm{Ortega}, \binits{P.A.}},
\oauthor{\bsnm{Kunesch}, \binits{M.}},
\oauthor{\bsnm{Del{\'e}tang}, \binits{G.}},
\oauthor{\bsnm{Genewein}, \binits{T.}},
\oauthor{\bsnm{Grau-Moya}, \binits{J.}},
\oauthor{\bsnm{Veness}, \binits{J.}},
\oauthor{\bsnm{Buchli}, \binits{J.}},
\oauthor{\bsnm{Degrave}, \binits{J.}},
\oauthor{\bsnm{Piot}, \binits{B.}},
\oauthor{\bsnm{Perolat}, \binits{J.}}, et al.:
Shaking the foundations: delusions in sequence models for interaction and control.
arXiv preprint arXiv:2110.10819
(2021)
\end{botherref}
\endbibitem

\bibitem[\protect\citeauthoryear{Niu et~al.}{2021}]{niu2021counterfactual}
\begin{bchapter}
\bauthor{\bsnm{Niu}, \binits{Y.}},
\bauthor{\bsnm{Tang}, \binits{K.}},
\bauthor{\bsnm{Zhang}, \binits{H.}},
\bauthor{\bsnm{Lu}, \binits{Z.}},
\bauthor{\bsnm{Hua}, \binits{X.-S.}},
\bauthor{\bsnm{Wen}, \binits{J.-R.}}:
\bctitle{Counterfactual {VQA}: A cause-effect look at language bias}.
In: \bbtitle{Proceedings of the IEEE/CVF Conference on Computer Vision and Pattern Recognition},
pp. \bfpage{12700}--\blpage{12710}
(\byear{2021})
\end{bchapter}
\endbibitem

\bibitem[\protect\citeauthoryear{Alzantot et~al.}{2018}]{alzantot2018generating}
\begin{bchapter}
\bauthor{\bsnm{Alzantot}, \binits{M.}},
\bauthor{\bsnm{Sharma}, \binits{Y.}},
\bauthor{\bsnm{Elgohary}, \binits{A.}},
\bauthor{\bsnm{Ho}, \binits{B.-J.}},
\bauthor{\bsnm{Srivastava}, \binits{M.}},
\bauthor{\bsnm{Chang}, \binits{K.-W.}}:
\bctitle{Generating natural language adversarial examples}.
In: \beditor{\bsnm{Riloff}, \binits{E.}},
\beditor{\bsnm{Chiang}, \binits{D.}},
\beditor{\bsnm{Hockenmaier}, \binits{J.}},
\beditor{\bsnm{Tsujii}, \binits{J.}} (eds.)
\bbtitle{Proceedings of the 2018 Conference on Empirical Methods in Natural Language Processing},
pp. \bfpage{2890}--\blpage{2896}.
\bpublisher{Association for Computational Linguistics},
\blocation{Brussels, Belgium}
(\byear{2018}).
\doiurl{10.18653/v1/D18-1316}
\end{bchapter}
\endbibitem

\bibitem[\protect\citeauthoryear{Chen et~al.}{2022}]{chen2022invariance}
\begin{bchapter}
\bauthor{\bsnm{Chen}, \binits{Y.}},
\bauthor{\bsnm{Zhang}, \binits{Y.}},
\bauthor{\bsnm{Bian}, \binits{Y.}},
\bauthor{\bsnm{Yang}, \binits{H.}},
\bauthor{\bsnm{KAILI}, \binits{M.}},
\bauthor{\bsnm{Xie}, \binits{B.}},
\bauthor{\bsnm{Liu}, \binits{T.}},
\bauthor{\bsnm{Han}, \binits{B.}},
\bauthor{\bsnm{Cheng}, \binits{J.}}:
\bctitle{Invariance principle meets out-of-distribution generalization on graphs}.
In: \bbtitle{ICML 2022: Workshop on Spurious Correlations, Invariance and Stability}
(\byear{2022})
\end{bchapter}
\endbibitem

\bibitem[\protect\citeauthoryear{Feng et~al.}{2021}]{feng2021should}
\begin{bchapter}
\bauthor{\bsnm{Feng}, \binits{F.}},
\bauthor{\bsnm{Huang}, \binits{W.}},
\bauthor{\bsnm{He}, \binits{X.}},
\bauthor{\bsnm{Xin}, \binits{X.}},
\bauthor{\bsnm{Wang}, \binits{Q.}},
\bauthor{\bsnm{Chua}, \binits{T.-S.}}:
\bctitle{Should graph convolution trust neighbors? a simple causal inference method}.
In: \bbtitle{Proceedings of the 44th International ACM SIGIR Conference on Research and Development in Information Retrieval}.
\bsertitle{SIGIR '21},
pp. \bfpage{1208}--\blpage{1218}.
\bpublisher{Association for Computing Machinery},
\blocation{New York, NY, USA}
(\byear{2021}).
\doiurl{10.1145/3404835.3462971}
\end{bchapter}
\endbibitem

\bibitem[\protect\citeauthoryear{Pearl}{2009}]{pearl2000models}
\begin{bbook}
\bauthor{\bsnm{Pearl}, \binits{J.}}:
\bbtitle{Causality: Models, Reasoning, and Inference},
\bedition{2}nd edn.
\bpublisher{Cambridge University Press},
\blocation{Cambridge}
(\byear{2009})
\end{bbook}
\endbibitem

\bibitem[\protect\citeauthoryear{Glymour et~al.}{2016}]{glymour2016causal}
\begin{botherref}
\oauthor{\bsnm{Glymour}, \binits{M.}},
\oauthor{\bsnm{Pearl}, \binits{J.}},
\oauthor{\bsnm{Jewell}, \binits{N.P.}}:
Causal inference in statistics: A primer.
John Wiley \& Sons
(2016)
\end{botherref}
\endbibitem

\bibitem[\protect\citeauthoryear{Peters et~al.}{2017}]{peters2017elements}
\begin{botherref}
\oauthor{\bsnm{Peters}, \binits{J.}},
\oauthor{\bsnm{Janzing}, \binits{D.}},
\oauthor{\bsnm{Sch{\"o}lkopf}, \binits{B.}}:
Elements of Causal Inference: Foundations and Learning Algorithms.
The MIT Press
(2017)
\end{botherref}
\endbibitem

\bibitem[\protect\citeauthoryear{Spirtes et~al.}{2000}]{spirtes2000causation}
\begin{botherref}
\oauthor{\bsnm{Spirtes}, \binits{P.}},
\oauthor{\bsnm{Glymour}, \binits{C.N.}},
\oauthor{\bsnm{Scheines}, \binits{R.}},
\oauthor{\bsnm{Heckerman}, \binits{D.}}:
Causation, prediction, and search.
MIT press
(2000)
\end{botherref}
\endbibitem

\bibitem[\protect\citeauthoryear{Sontakke et~al.}{2021}]{sontakke2021causal}
\begin{bchapter}
\bauthor{\bsnm{Sontakke}, \binits{S.A.}},
\bauthor{\bsnm{Mehrjou}, \binits{A.}},
\bauthor{\bsnm{Itti}, \binits{L.}},
\bauthor{\bsnm{Sch{\"o}lkopf}, \binits{B.}}:
\bctitle{Causal curiosity: {RL} agents discovering self-supervised experiments for causal representation learning}.
In: \bbtitle{International Conference on Machine Learning},
vol. \bseriesno{139},
pp. \bfpage{9848}--\blpage{9858}
(\byear{2021}).
\bcomment{PMLR}
\end{bchapter}
\endbibitem

\bibitem[\protect\citeauthoryear{Thompson et~al.}{2022}]{thompson2022identification}
\begin{barticle}
\bauthor{\bsnm{Thompson}, \binits{J.}},
\bauthor{\bsnm{{Kasun Prasanga}}, \binits{D.}},
\bauthor{\bsnm{Murakami}, \binits{T.}}:
\batitle{Identification of unknown object properties based on tactile motion sequence using 2-finger gripper robot}.
\bjtitle{Precision Engineering}
\bvolume{74},
\bfpage{347}--\blpage{357}
(\byear{2022})
\doiurl{10.1016/j.precisioneng.2021.12.009}
\end{barticle}
\endbibitem

\bibitem[\protect\citeauthoryear{MKC and Shimono}{2018}]{mkc2018inertia}
\begin{barticle}
\bauthor{\bsnm{MKC}, \binits{D.C.}},
\bauthor{\bsnm{Shimono}, \binits{T.}}:
\batitle{Inertia compensation of motion copying system for dexterous object handling}.
\bjtitle{IEEJ Journal of Industry Applications}
\bvolume{7}(\bissue{6}),
\bfpage{495}--\blpage{505}
(\byear{2018})
\doiurl{10.1541/ieejjia.7.495}
\end{barticle}
\endbibitem

\bibitem[\protect\citeauthoryear{Murali et~al.}{2020}]{murali2018learning}
\begin{bchapter}
\bauthor{\bsnm{Murali}, \binits{A.}},
\bauthor{\bsnm{Li}, \binits{Y.}},
\bauthor{\bsnm{Gandhi}, \binits{D.}},
\bauthor{\bsnm{Gupta}, \binits{A.}}:
\bctitle{Learning to grasp without seeing}.
In: \beditor{\bsnm{Xiao}, \binits{J.}},
\beditor{\bsnm{Kr{\"o}ger}, \binits{T.}},
\beditor{\bsnm{Khatib}, \binits{O.}} (eds.)
\bbtitle{Proceedings of the 2018 International Symposium on Experimental Robotics},
pp. \bfpage{375}--\blpage{386}.
\bpublisher{Springer},
\blocation{Cham}
(\byear{2020})
\end{bchapter}
\endbibitem

\bibitem[\protect\citeauthoryear{Yuan et~al.}{2017}]{yuan2017shape}
\begin{bchapter}
\bauthor{\bsnm{Yuan}, \binits{W.}},
\bauthor{\bsnm{Zhu}, \binits{C.}},
\bauthor{\bsnm{Owens}, \binits{A.}},
\bauthor{\bsnm{Srinivasan}, \binits{M.A.}},
\bauthor{\bsnm{Adelson}, \binits{E.H.}}:
\bctitle{Shape-independent hardness estimation using deep learning and a gelsight tactile sensor}.
In: \bbtitle{2017 IEEE International Conference on Robotics and Automation (ICRA)},
pp. \bfpage{951}--\blpage{958}
(\byear{2017}).
\doiurl{10.1109/ICRA.2017.7989116}
\end{bchapter}
\endbibitem

\bibitem[\protect\citeauthoryear{Yi et~al.}{2016}]{yi2016active}
\begin{bchapter}
\bauthor{\bsnm{Yi}, \binits{Z.}},
\bauthor{\bsnm{Calandra}, \binits{R.}},
\bauthor{\bsnm{Veiga}, \binits{F.}},
\bauthor{\bsnm{Hoof}, \binits{H.}},
\bauthor{\bsnm{Hermans}, \binits{T.}},
\bauthor{\bsnm{Zhang}, \binits{Y.}},
\bauthor{\bsnm{Peters}, \binits{J.}}:
\bctitle{Active tactile object exploration with gaussian processes}.
In: \bbtitle{2016 IEEE/RSJ International Conference on Intelligent Robots and Systems (IROS)},
pp. \bfpage{4925}--\blpage{4930}
(\byear{2016}).
\doiurl{10.1109/IROS.2016.7759723}
\end{bchapter}
\endbibitem

\bibitem[\protect\citeauthoryear{Kaboli et~al.}{2017}]{kaboli2017tactile}
\begin{barticle}
\bauthor{\bsnm{Kaboli}, \binits{M.}},
\bauthor{\bsnm{Feng}, \binits{D.}},
\bauthor{\bsnm{Yao}, \binits{K.}},
\bauthor{\bsnm{Lanillos}, \binits{P.}},
\bauthor{\bsnm{Cheng}, \binits{G.}}:
\batitle{A tactile-based framework for active object learning and discrimination using multimodal robotic skin}.
\bjtitle{IEEE Robotics and Automation Letters}
\bvolume{2}(\bissue{4}),
\bfpage{2143}--\blpage{2150}
(\byear{2017})
\doiurl{10.1109/LRA.2017.2720853}
\end{barticle}
\endbibitem

\bibitem[\protect\citeauthoryear{Yu et~al.}{2017}]{yu2017preparing}
\begin{botherref}
\oauthor{\bsnm{Yu}, \binits{W.}},
\oauthor{\bsnm{Tan}, \binits{J.}},
\oauthor{\bsnm{Liu}, \binits{C.K.}},
\oauthor{\bsnm{Turk}, \binits{G.}}:
Preparing for the unknown: Learning a universal policy with online system identification.
arXiv preprint arXiv:1702.02453
(2017)
\end{botherref}
\endbibitem

\bibitem[\protect\citeauthoryear{Murooka et~al.}{2017}]{murooka2017feasibility}
\begin{bchapter}
\bauthor{\bsnm{Murooka}, \binits{M.}},
\bauthor{\bsnm{Nozawa}, \binits{S.}},
\bauthor{\bsnm{Kakiuchi}, \binits{Y.}},
\bauthor{\bsnm{Okada}, \binits{K.}},
\bauthor{\bsnm{Inaba}, \binits{M.}}:
\bctitle{Feasibility evaluation of object manipulation by a humanoid robot based on recursive estimation of the object's physical properties}.
In: \bbtitle{2017 IEEE International Conference on Robotics and Automation (ICRA)},
pp. \bfpage{4082}--\blpage{4089}
(\byear{2017}).
\doiurl{10.1109/ICRA.2017.7989469}
\end{bchapter}
\endbibitem

\bibitem[\protect\citeauthoryear{Mavrakis and Stolkin}{2020}]{mavrakis2020estimation}
\begin{barticle}
\bauthor{\bsnm{Mavrakis}, \binits{N.}},
\bauthor{\bsnm{Stolkin}, \binits{R.}}:
\batitle{Estimation and exploitation of objects’ inertial parameters in robotic grasping and manipulation: A survey}.
\bjtitle{Robotics and Autonomous Systems}
\bvolume{124},
\bfpage{103374}
(\byear{2020})
\doiurl{10.1016/j.robot.2019.103374}
\end{barticle}
\endbibitem

\bibitem[\protect\citeauthoryear{Schulman et~al.}{2017}]{schulman2017proximal}
\begin{botherref}
\oauthor{\bsnm{Schulman}, \binits{J.}},
\oauthor{\bsnm{Wolski}, \binits{F.}},
\oauthor{\bsnm{Dhariwal}, \binits{P.}},
\oauthor{\bsnm{Radford}, \binits{A.}},
\oauthor{\bsnm{Klimov}, \binits{O.}}:
Proximal policy optimization algorithms.
arXiv preprint arXiv:1707.06347
(2017)
\end{botherref}
\endbibitem

\bibitem[\protect\citeauthoryear{De~Boer et~al.}{2005}]{de2005tutorial}
\begin{barticle}
\bauthor{\bsnm{De~Boer}, \binits{P.-T.}},
\bauthor{\bsnm{Kroese}, \binits{D.P.}},
\bauthor{\bsnm{Mannor}, \binits{S.}},
\bauthor{\bsnm{Rubinstein}, \binits{R.Y.}}:
\batitle{A tutorial on the cross-entropy method}.
\bjtitle{Annals of operations research}
\bvolume{134},
\bfpage{19}--\blpage{67}
(\byear{2005})
\end{barticle}
\endbibitem

\bibitem[\protect\citeauthoryear{Camacho and Alba}{2013}]{camacho2013model}
\begin{botherref}
\oauthor{\bsnm{Camacho}, \binits{E.F.}},
\oauthor{\bsnm{Alba}, \binits{C.B.}}:
Model predictive control.
Springer
(2013)
\end{botherref}
\endbibitem

\bibitem[\protect\citeauthoryear{Kaddour et~al.}{2022}]{kaddour2022causal}
\begin{botherref}
\oauthor{\bsnm{Kaddour}, \binits{J.}},
\oauthor{\bsnm{Lynch}, \binits{A.}},
\oauthor{\bsnm{Liu}, \binits{Q.}},
\oauthor{\bsnm{Kusner}, \binits{M.J.}},
\oauthor{\bsnm{Silva}, \binits{R.}}:
Causal machine learning: A survey and open problems.
arXiv preprint arXiv:2206.15475
(2022)
\end{botherref}
\endbibitem

\bibitem[\protect\citeauthoryear{Bareinboim}{2020}]{bareinboimcrlonline}
\begin{botherref}
\oauthor{\bsnm{Bareinboim}, \binits{E.}}:
Towards Causal reinforcement learning, ICML tutorial
(2020).
\url{https://crl.causalai.net/}
\end{botherref}
\endbibitem

\bibitem[\protect\citeauthoryear{Zeng et~al.}{2024}]{zeng2023survey}
\begin{botherref}
\oauthor{\bsnm{Zeng}, \binits{Y.}},
\oauthor{\bsnm{Cai}, \binits{R.}},
\oauthor{\bsnm{Sun}, \binits{F.}},
\oauthor{\bsnm{Huang}, \binits{L.}},
\oauthor{\bsnm{Hao}, \binits{Z.}}:
A survey on causal reinforcement learning.
IEEE Transactions on Neural Networks and Learning Systems,
1--21
(2024)
\doiurl{10.1109/TNNLS.2024.3403001}
\end{botherref}
\endbibitem

\bibitem[\protect\citeauthoryear{Grimbly et~al.}{2021}]{grimbly2021causal}
\begin{botherref}
\oauthor{\bsnm{Grimbly}, \binits{S.J.}},
\oauthor{\bsnm{Shock}, \binits{J.}},
\oauthor{\bsnm{Pretorius}, \binits{A.}}:
Causal multi-agent reinforcement learning: Review and open problems.
arXiv preprint arXiv:2111.06721
(2021)
\end{botherref}
\endbibitem

\bibitem[\protect\citeauthoryear{Weichwald et~al.}{2022}]{weichwald2022learning}
\begin{bchapter}
\bauthor{\bsnm{Weichwald}, \binits{S.}},
\bauthor{\bsnm{Mogensen}, \binits{S.W.}},
\bauthor{\bsnm{Lee}, \binits{T.E.}},
\bauthor{\bsnm{Baumann}, \binits{D.}},
\bauthor{\bsnm{Kroemer}, \binits{O.}},
\bauthor{\bsnm{Guyon}, \binits{I.}},
\bauthor{\bsnm{Trimpe}, \binits{S.}},
\bauthor{\bsnm{Peters}, \binits{J.}},
\bauthor{\bsnm{Pfister}, \binits{N.}}:
\bctitle{Learning by doing: Controlling a dynamical system using causality, control, and reinforcement learning}.
In: \bbtitle{NeurIPS 2021 Competitions and Demonstrations Track},
vol. \bseriesno{176},
pp. \bfpage{246}--\blpage{258}
(\byear{2022}).
\bcomment{PMLR}
\end{bchapter}
\endbibitem

\bibitem[\protect\citeauthoryear{Bareinboim et~al.}{2015}]{bareinboim2015bandits}
\begin{botherref}
\oauthor{\bsnm{Bareinboim}, \binits{E.}},
\oauthor{\bsnm{Forney}, \binits{A.}},
\oauthor{\bsnm{Pearl}, \binits{J.}}:
Bandits with unobserved confounders: A causal approach.
Advances in Neural Information Processing Systems
\textbf{28}
(2015)
\end{botherref}
\endbibitem

\bibitem[\protect\citeauthoryear{Gershman}{2017}]{gershman2017reinforcement}
\begin{barticle}
\bauthor{\bsnm{Gershman}, \binits{S.J.}}:
\batitle{Reinforcement learning and causal models}.
\bjtitle{The Oxford handbook of causal reasoning}
\bvolume{1},
\bfpage{295}
(\byear{2017})
\end{barticle}
\endbibitem

\bibitem[\protect\citeauthoryear{Dasgupta et~al.}{2019}]{dasgupta2019causal}
\begin{botherref}
\oauthor{\bsnm{Dasgupta}, \binits{I.}},
\oauthor{\bsnm{Wang}, \binits{J.}},
\oauthor{\bsnm{Chiappa}, \binits{S.}},
\oauthor{\bsnm{Mitrovic}, \binits{J.}},
\oauthor{\bsnm{Ortega}, \binits{P.}},
\oauthor{\bsnm{Raposo}, \binits{D.}},
\oauthor{\bsnm{Hughes}, \binits{E.}},
\oauthor{\bsnm{Battaglia}, \binits{P.}},
\oauthor{\bsnm{Botvinick}, \binits{M.}},
\oauthor{\bsnm{Kurth-Nelson}, \binits{Z.}}:
Causal reasoning from meta-reinforcement learning.
arXiv preprint arXiv:1901.08162
(2019)
\end{botherref}
\endbibitem

\bibitem[\protect\citeauthoryear{Amin et~al.}{2021}]{amin2021survey}
\begin{botherref}
\oauthor{\bsnm{Amin}, \binits{S.}},
\oauthor{\bsnm{Gomrokchi}, \binits{M.}},
\oauthor{\bsnm{Satija}, \binits{H.}},
\oauthor{\bsnm{Hoof}, \binits{H.}},
\oauthor{\bsnm{Precup}, \binits{D.}}:
A survey of exploration methods in reinforcement learning.
arXiv preprint arXiv:2109.00157
(2021)
\end{botherref}
\endbibitem

\bibitem[\protect\citeauthoryear{Peng et~al.}{2022}]{peng2022causality}
\begin{barticle}
\bauthor{\bsnm{Peng}, \binits{S.}},
\bauthor{\bsnm{Hu}, \binits{X.}},
\bauthor{\bsnm{Zhang}, \binits{R.}},
\bauthor{\bsnm{Tang}, \binits{K.}},
\bauthor{\bsnm{Guo}, \binits{J.}},
\bauthor{\bsnm{Yi}, \binits{Q.}},
\bauthor{\bsnm{Chen}, \binits{R.}},
\bauthor{\bsnm{Zhang}, \binits{X.}},
\bauthor{\bsnm{Du}, \binits{Z.}},
\bauthor{\bsnm{Li}, \binits{L.}},
\bauthor{\bsnm{Guo}, \binits{Q.}},
\bauthor{\bsnm{Chen}, \binits{Y.}}:
\batitle{Causality-driven hierarchical structure discovery for reinforcement learning}.
\bjtitle{Advances in Neural Information Processing Systems}
\bvolume{35},
\bfpage{20064}--\blpage{20076}
(\byear{2022})
\end{barticle}
\endbibitem

\bibitem[\protect\citeauthoryear{Rezende et~al.}{2020}]{rezende2020causally}
\begin{botherref}
\oauthor{\bsnm{Rezende}, \binits{D.J.}},
\oauthor{\bsnm{Danihelka}, \binits{I.}},
\oauthor{\bsnm{Papamakarios}, \binits{G.}},
\oauthor{\bsnm{Ke}, \binits{N.R.}},
\oauthor{\bsnm{Jiang}, \binits{R.}},
\oauthor{\bsnm{Weber}, \binits{T.}},
\oauthor{\bsnm{Gregor}, \binits{K.}},
\oauthor{\bsnm{Merzic}, \binits{H.}},
\oauthor{\bsnm{Viola}, \binits{F.}},
\oauthor{\bsnm{Wang}, \binits{J.}}, et al.:
Causally correct partial models for reinforcement learning.
arXiv preprint arXiv:2002.02836
(2020)
\end{botherref}
\endbibitem

\bibitem[\protect\citeauthoryear{Pitis et~al.}{2020}]{pitis2020counterfactual}
\begin{barticle}
\bauthor{\bsnm{Pitis}, \binits{S.}},
\bauthor{\bsnm{Creager}, \binits{E.}},
\bauthor{\bsnm{Garg}, \binits{A.}}:
\batitle{Counterfactual data augmentation using locally factored dynamics}.
\bjtitle{Advances in Neural Information Processing Systems}
\bvolume{33},
\bfpage{3976}--\blpage{3990}
(\byear{2020})
\end{barticle}
\endbibitem

\bibitem[\protect\citeauthoryear{Seitzer et~al.}{2021}]{seitzer2021causal}
\begin{barticle}
\bauthor{\bsnm{Seitzer}, \binits{M.}},
\bauthor{\bsnm{Sch{\"o}lkopf}, \binits{B.}},
\bauthor{\bsnm{Martius}, \binits{G.}}:
\batitle{Causal influence detection for improving efficiency in reinforcement learning}.
\bjtitle{Advances in Neural Information Processing Systems}
\bvolume{34},
\bfpage{22905}--\blpage{22918}
(\byear{2021})
\end{barticle}
\endbibitem

\bibitem[\protect\citeauthoryear{Molina et~al.}{2020}]{molina2020causal}
\begin{barticle}
\bauthor{\bsnm{Molina}, \binits{A.M.}},
\bauthor{\bsnm{Avelino}, \binits{I.F.}},
\bauthor{\bsnm{Morales}, \binits{E.F.}},
\bauthor{\bsnm{Sucar}, \binits{L.E.}}:
\batitle{Causal based {Q}-learning}.
\bjtitle{Research in Computing Science}
\bvolume{149},
\bfpage{95}--\blpage{104}
(\byear{2020})
\end{barticle}
\endbibitem

\bibitem[\protect\citeauthoryear{Pathak et~al.}{2017}]{pathak2017curiosity}
\begin{bchapter}
\bauthor{\bsnm{Pathak}, \binits{D.}},
\bauthor{\bsnm{Agrawal}, \binits{P.}},
\bauthor{\bsnm{Efros}, \binits{A.A.}},
\bauthor{\bsnm{Darrell}, \binits{T.}}:
\bctitle{Curiosity-driven exploration by self-supervised prediction}.
In: \bbtitle{International Conference on Machine Learning},
vol. \bseriesno{70},
pp. \bfpage{2778}--\blpage{2787}
(\byear{2017}).
\bcomment{PMLR}
\end{bchapter}
\endbibitem

\bibitem[\protect\citeauthoryear{Burda et~al.}{2019}]{burda2018large}
\begin{bchapter}
\bauthor{\bsnm{Burda}, \binits{Y.}},
\bauthor{\bsnm{Edwards}, \binits{H.}},
\bauthor{\bsnm{Pathak}, \binits{D.}},
\bauthor{\bsnm{Storkey}, \binits{A.}},
\bauthor{\bsnm{Darrell}, \binits{T.}},
\bauthor{\bsnm{Efros}, \binits{A.A.}}:
\bctitle{Large-scale study of curiosity-driven learning}.
In: \bbtitle{International Conference on Learning Representations}
(\byear{2019})
\end{bchapter}
\endbibitem

\bibitem[\protect\citeauthoryear{Schmidhuber}{1991}]{schmidhuber1991curious}
\begin{bchapter}
\bauthor{\bsnm{Schmidhuber}, \binits{J.}}:
\bctitle{Curious model-building control systems}.
In: \bbtitle{Proceedings 1991 IEEE International Joint Conference on Neural Networks},
pp. \bfpage{1458}--\blpage{14632}
(\byear{1991}).
\doiurl{10.1109/IJCNN.1991.170605}
\end{bchapter}
\endbibitem

\bibitem[\protect\citeauthoryear{Chentanez et~al.}{2004}]{chentanez2004intrinsically}
\begin{botherref}
\oauthor{\bsnm{Chentanez}, \binits{N.}},
\oauthor{\bsnm{Barto}, \binits{A.}},
\oauthor{\bsnm{Singh}, \binits{S.}}:
Intrinsically motivated reinforcement learning.
Advances in neural information processing systems
\textbf{17}
(2004)
\end{botherref}
\endbibitem

\bibitem[\protect\citeauthoryear{Lehman et~al.}{2008}]{lehman2008exploiting}
\begin{bchapter}
\bauthor{\bsnm{Lehman}, \binits{J.}},
\bauthor{\bsnm{Stanley}, \binits{K.O.}}, \betal:
\bctitle{Exploiting open-endedness to solve problems through the search for novelty.}
In: \bbtitle{ALIFE},
pp. \bfpage{329}--\blpage{336}
(\byear{2008})
\end{bchapter}
\endbibitem

\bibitem[\protect\citeauthoryear{Oudeyer and Kaplan}{2007}]{oudeyer2009intrinsic}
\begin{botherref}
\oauthor{\bsnm{Oudeyer}, \binits{P.-Y.}},
\oauthor{\bsnm{Kaplan}, \binits{F.}}:
What is intrinsic motivation? a typology of computational approaches.
Frontiers in Neurorobotics
\textbf{1}
(2007)
\doiurl{10.3389/neuro.12.006.2007}
\end{botherref}
\endbibitem

\bibitem[\protect\citeauthoryear{Sun et~al.}{2011}]{sun2011planning}
\begin{bchapter}
\bauthor{\bsnm{Sun}, \binits{Y.}},
\bauthor{\bsnm{Gomez}, \binits{F.}},
\bauthor{\bsnm{Schmidhuber}, \binits{J.}}:
\bctitle{Planning to be surprised: Optimal bayesian exploration in dynamic environments}.
In: \beditor{\bsnm{Schmidhuber}, \binits{J.}},
\beditor{\bsnm{Th{\'o}risson}, \binits{K.R.}},
\beditor{\bsnm{Looks}, \binits{M.}} (eds.)
\bbtitle{Artificial General Intelligence},
pp. \bfpage{41}--\blpage{51}.
\bpublisher{Springer},
\blocation{Berlin, Heidelberg}
(\byear{2011})
\end{bchapter}
\endbibitem

\bibitem[\protect\citeauthoryear{Still and Precup}{2012}]{still2012information}
\begin{barticle}
\bauthor{\bsnm{Still}, \binits{S.}},
\bauthor{\bsnm{Precup}, \binits{D.}}:
\batitle{An information-theoretic approach to curiosity-driven reinforcement learning}.
\bjtitle{Theory in Biosciences}
\bvolume{131},
\bfpage{139}--\blpage{148}
(\byear{2012})
\end{barticle}
\endbibitem

\bibitem[\protect\citeauthoryear{Baldassarre and Mirolli}{2013}]{baldassarre2013intrinsically}
\begin{botherref}
\oauthor{\bsnm{Baldassarre}, \binits{G.}},
\oauthor{\bsnm{Mirolli}, \binits{M.}}:
Intrinsically motivated learning in natural and artificial systems.
Springer
(2013)
\end{botherref}
\endbibitem

\bibitem[\protect\citeauthoryear{Baranes and Oudeyer}{2013}]{baranes2013active}
\begin{barticle}
\bauthor{\bsnm{Baranes}, \binits{A.}},
\bauthor{\bsnm{Oudeyer}, \binits{P.-Y.}}:
\batitle{Active learning of inverse models with intrinsically motivated goal exploration in robots}.
\bjtitle{Robotics and Autonomous Systems}
\bvolume{61}(\bissue{1}),
\bfpage{49}--\blpage{73}
(\byear{2013})
\doiurl{10.1016/j.robot.2012.05.008}
\end{barticle}
\endbibitem

\bibitem[\protect\citeauthoryear{Barto}{2013}]{barto2013intrinsic}
\begin{bbook}
\bauthor{\bsnm{Barto}, \binits{A.G.}}:
In: \beditor{\bsnm{Baldassarre}, \binits{G.}},
\beditor{\bsnm{Mirolli}, \binits{M.}} (eds.)
\bbtitle{Intrinsic Motivation and Reinforcement Learning},
pp. \bfpage{17}--\blpage{47}.
\bpublisher{Springer},
\blocation{Berlin, Heidelberg}
(\byear{2013}).
\doiurl{10.1007/978-3-642-32375-1_2}
\end{bbook}
\endbibitem

\bibitem[\protect\citeauthoryear{Stadie et~al.}{2015}]{stadie2015incentivizing}
\begin{botherref}
\oauthor{\bsnm{Stadie}, \binits{B.C.}},
\oauthor{\bsnm{Levine}, \binits{S.}},
\oauthor{\bsnm{Abbeel}, \binits{P.}}:
Incentivizing exploration in reinforcement learning with deep predictive models.
arXiv preprint arXiv:1507.00814
(2015)
\end{botherref}
\endbibitem

\bibitem[\protect\citeauthoryear{Mohamed and Jimenez~Rezende}{2015}]{mohamed2015variational}
\begin{botherref}
\oauthor{\bsnm{Mohamed}, \binits{S.}},
\oauthor{\bsnm{Jimenez~Rezende}, \binits{D.}}:
Variational information maximisation for intrinsically motivated reinforcement learning.
Advances in neural information processing systems
\textbf{28}
(2015)
\end{botherref}
\endbibitem

\bibitem[\protect\citeauthoryear{Houthooft et~al.}{2016}]{houthooft2016vime}
\begin{botherref}
\oauthor{\bsnm{Houthooft}, \binits{R.}},
\oauthor{\bsnm{Chen}, \binits{X.}},
\oauthor{\bsnm{Duan}, \binits{Y.}},
\oauthor{\bsnm{Schulman}, \binits{J.}},
\oauthor{\bsnm{De~Turck}, \binits{F.}},
\oauthor{\bsnm{Abbeel}, \binits{P.}}:
{VIME}: Variational information maximizing exploration.
Advances in neural information processing systems
\textbf{29}
(2016)
\end{botherref}
\endbibitem

\bibitem[\protect\citeauthoryear{Osband et~al.}{2016}]{osband2016deep}
\begin{botherref}
\oauthor{\bsnm{Osband}, \binits{I.}},
\oauthor{\bsnm{Blundell}, \binits{C.}},
\oauthor{\bsnm{Pritzel}, \binits{A.}},
\oauthor{\bsnm{Van~Roy}, \binits{B.}}:
Deep exploration via bootstrapped {DQN}.
Advances in neural information processing systems
\textbf{29}
(2016)
\end{botherref}
\endbibitem

\bibitem[\protect\citeauthoryear{Forestier et~al.}{2022}]{forestier2017intrinsically}
\begin{barticle}
\bauthor{\bsnm{Forestier}, \binits{S.}},
\bauthor{\bsnm{Portelas}, \binits{R.}},
\bauthor{\bsnm{Mollard}, \binits{Y.}},
\bauthor{\bsnm{Oudeyer}, \binits{P.-Y.}}:
\batitle{Intrinsically motivated goal exploration processes with automatic curriculum learning}.
\bjtitle{Journal of Machine Learning Research}
\bvolume{23}(\bissue{152}),
\bfpage{1}--\blpage{41}
(\byear{2022})
\end{barticle}
\endbibitem

\bibitem[\protect\citeauthoryear{Tang et~al.}{2017}]{tang2017exploration}
\begin{botherref}
\oauthor{\bsnm{Tang}, \binits{H.}},
\oauthor{\bsnm{Houthooft}, \binits{R.}},
\oauthor{\bsnm{Foote}, \binits{D.}},
\oauthor{\bsnm{Stooke}, \binits{A.}},
\oauthor{\bsnm{Xi~Chen}, \binits{O.}},
\oauthor{\bsnm{Duan}, \binits{Y.}},
\oauthor{\bsnm{Schulman}, \binits{J.}},
\oauthor{\bsnm{DeTurck}, \binits{F.}},
\oauthor{\bsnm{Abbeel}, \binits{P.}}:
\#exploration: A study of count-based exploration for deep reinforcement learning.
Advances in neural information processing systems
\textbf{30}
(2017)
\end{botherref}
\endbibitem

\bibitem[\protect\citeauthoryear{Colas et~al.}{2018}]{colas2018gep}
\begin{bchapter}
\bauthor{\bsnm{Colas}, \binits{C.}},
\bauthor{\bsnm{Sigaud}, \binits{O.}},
\bauthor{\bsnm{Oudeyer}, \binits{P.-Y.}}:
\bctitle{Gep-pg: Decoupling exploration and exploitation in deep reinforcement learning algorithms}.
In: \bbtitle{International Conference on Machine Learning},
vol. \bseriesno{80},
pp. \bfpage{1039}--\blpage{1048}
(\byear{2018}).
\bcomment{PMLR}
\end{bchapter}
\endbibitem

\bibitem[\protect\citeauthoryear{Laversanne-Finot et~al.}{2018}]{laversanne2018curiosity}
\begin{bchapter}
\bauthor{\bsnm{Laversanne-Finot}, \binits{A.}},
\bauthor{\bsnm{Pere}, \binits{A.}},
\bauthor{\bsnm{Oudeyer}, \binits{P.-Y.}}:
\bctitle{Curiosity driven exploration of learned disentangled goal spaces}.
In: \bbtitle{Conference on Robot Learning},
vol. \bseriesno{87},
pp. \bfpage{487}--\blpage{504}
(\byear{2018}).
\bcomment{PMLR}
\end{bchapter}
\endbibitem

\bibitem[\protect\citeauthoryear{Oudeyer}{2018}]{oudeyer2018computational}
\begin{botherref}
\oauthor{\bsnm{Oudeyer}, \binits{P.-Y.}}:
Computational theories of curiosity-driven learning.
arXiv preprint arXiv:1802.10546
(2018)
\end{botherref}
\endbibitem

\bibitem[\protect\citeauthoryear{Guo et~al.}{2022}]{guo2022byol}
\begin{barticle}
\bauthor{\bsnm{Guo}, \binits{Z.}},
\bauthor{\bsnm{Thakoor}, \binits{S.}},
\bauthor{\bsnm{P{\^\i}slar}, \binits{M.}},
\bauthor{\bsnm{Avila~Pires}, \binits{B.}},
\bauthor{\bsnm{Altch{\'e}}, \binits{F.}},
\bauthor{\bsnm{Tallec}, \binits{C.}},
\bauthor{\bsnm{Saade}, \binits{A.}},
\bauthor{\bsnm{Calandriello}, \binits{D.}},
\bauthor{\bsnm{Grill}, \binits{J.-B.}},
\bauthor{\bsnm{Tang}, \binits{Y.}}, \betal:
\batitle{{BYOL}-{E}xplore: Exploration by bootstrapped prediction}.
\bjtitle{Advances in neural information processing systems}
\bvolume{35},
\bfpage{31855}--\blpage{31870}
(\byear{2022})
\end{barticle}
\endbibitem

\bibitem[\protect\citeauthoryear{Gasse et~al.}{2021}]{gasse2021causal}
\begin{botherref}
\oauthor{\bsnm{Gasse}, \binits{M.}},
\oauthor{\bsnm{Grasset}, \binits{D.}},
\oauthor{\bsnm{Gaudron}, \binits{G.}},
\oauthor{\bsnm{Oudeyer}, \binits{P.-Y.}}:
Causal reinforcement learning using observational and interventional data.
arXiv preprint arXiv:2106.14421
(2021)
\end{botherref}
\endbibitem

\bibitem[\protect\citeauthoryear{Ke et~al.}{2021}]{ke2021systematic}
\begin{bchapter}
\bauthor{\bsnm{Ke}, \binits{N.R.}},
\bauthor{\bsnm{Didolkar}, \binits{A.R.}},
\bauthor{\bsnm{Mittal}, \binits{S.}},
\bauthor{\bsnm{Goyal}, \binits{A.}},
\bauthor{\bsnm{Lajoie}, \binits{G.}},
\bauthor{\bsnm{Bauer}, \binits{S.}},
\bauthor{\bsnm{Rezende}, \binits{D.J.}},
\bauthor{\bsnm{Bengio}, \binits{Y.}},
\bauthor{\bsnm{Pal}, \binits{C.}},
\bauthor{\bsnm{Mozer}, \binits{M.C.}}:
\bctitle{Systematic evaluation of causal discovery in visual model based reinforcement learning}.
In: \bbtitle{Thirty-fifth Conference on Neural Information Processing Systems Datasets and Benchmarks Track (Round 2)}
(\byear{2021})
\end{bchapter}
\endbibitem

\bibitem[\protect\citeauthoryear{Nair et~al.}{2019}]{nair2019causal}
\begin{botherref}
\oauthor{\bsnm{Nair}, \binits{S.}},
\oauthor{\bsnm{Zhu}, \binits{Y.}},
\oauthor{\bsnm{Savarese}, \binits{S.}},
\oauthor{\bsnm{Fei-Fei}, \binits{L.}}:
Causal induction from visual observations for goal directed tasks.
arXiv preprint arXiv:1910.01751
(2019)
\end{botherref}
\endbibitem

\bibitem[\protect\citeauthoryear{Ding et~al.}{2022}]{ding2022generalizing}
\begin{barticle}
\bauthor{\bsnm{Ding}, \binits{W.}},
\bauthor{\bsnm{Lin}, \binits{H.}},
\bauthor{\bsnm{Li}, \binits{B.}},
\bauthor{\bsnm{Zhao}, \binits{D.}}:
\batitle{Generalizing goal-conditioned reinforcement learning with variational causal reasoning}.
\bjtitle{Advances in Neural Information Processing Systems}
\bvolume{35},
\bfpage{26532}--\blpage{26548}
(\byear{2022})
\end{barticle}
\endbibitem

\bibitem[\protect\citeauthoryear{Thomas et~al.}{2017}]{thomas2017independently}
\begin{botherref}
\oauthor{\bsnm{Thomas}, \binits{V.}},
\oauthor{\bsnm{Pondard}, \binits{J.}},
\oauthor{\bsnm{Bengio}, \binits{E.}},
\oauthor{\bsnm{Sarfati}, \binits{M.}},
\oauthor{\bsnm{Beaudoin}, \binits{P.}},
\oauthor{\bsnm{Meurs}, \binits{M.-J.}},
\oauthor{\bsnm{Pineau}, \binits{J.}},
\oauthor{\bsnm{Precup}, \binits{D.}},
\oauthor{\bsnm{Bengio}, \binits{Y.}}:
Independently controllable factors.
arXiv preprint arXiv:1708.01289
(2017)
\end{botherref}
\endbibitem

\bibitem[\protect\citeauthoryear{Burgess et~al.}{2018}]{burgess2018understanding}
\begin{botherref}
\oauthor{\bsnm{Burgess}, \binits{C.P.}},
\oauthor{\bsnm{Higgins}, \binits{I.}},
\oauthor{\bsnm{Pal}, \binits{A.}},
\oauthor{\bsnm{Matthey}, \binits{L.}},
\oauthor{\bsnm{Watters}, \binits{N.}},
\oauthor{\bsnm{Desjardins}, \binits{G.}},
\oauthor{\bsnm{Lerchner}, \binits{A.}}:
Understanding disentangling in beta-{VAE}.
arXiv preprint arXiv:1804.03599
(2018)
\end{botherref}
\endbibitem

\bibitem[\protect\citeauthoryear{Kim and Mnih}{2018}]{kim2018disentangling}
\begin{bchapter}
\bauthor{\bsnm{Kim}, \binits{H.}},
\bauthor{\bsnm{Mnih}, \binits{A.}}:
\bctitle{Disentangling by factorising}.
In: \bbtitle{International Conference on Machine Learning},
vol. \bseriesno{80},
pp. \bfpage{2649}--\blpage{2658}
(\byear{2018}).
\bcomment{PMLR}
\end{bchapter}
\endbibitem

\bibitem[\protect\citeauthoryear{Chen et~al.}{2018}]{chen2018isolating}
\begin{botherref}
\oauthor{\bsnm{Chen}, \binits{R.T.}},
\oauthor{\bsnm{Li}, \binits{X.}},
\oauthor{\bsnm{Grosse}, \binits{R.B.}},
\oauthor{\bsnm{Duvenaud}, \binits{D.K.}}:
Isolating sources of disentanglement in variational autoencoders.
Advances in neural information processing systems
\textbf{31}
(2018)
\end{botherref}
\endbibitem

\bibitem[\protect\citeauthoryear{Volodin et~al.}{2020}]{volodin2020resolving}
\begin{botherref}
\oauthor{\bsnm{Volodin}, \binits{S.}},
\oauthor{\bsnm{Wichers}, \binits{N.}},
\oauthor{\bsnm{Nixon}, \binits{J.}}:
Resolving spurious correlations in causal models of environments via interventions.
arXiv preprint arXiv:2002.05217
(2020)
\end{botherref}
\endbibitem

\bibitem[\protect\citeauthoryear{Coumans et~al.}{2013}]{coumans2013bullet}
\begin{botherref}
\oauthor{\bsnm{Coumans}, \binits{E.}}, et al.:
Bullet real-time physics simulation.
URL http://bulletphysics.org
(2013)
\end{botherref}
\endbibitem

\bibitem[\protect\citeauthoryear{Wuthrich et~al.}{2021}]{wuthrich2021trifinger}
\begin{bchapter}
\bauthor{\bsnm{Wuthrich}, \binits{M.}},
\bauthor{\bsnm{Widmaier}, \binits{F.}},
\bauthor{\bsnm{Grimminger}, \binits{F.}},
\bauthor{\bsnm{Joshi}, \binits{S.}},
\bauthor{\bsnm{Agrawal}, \binits{V.}},
\bauthor{\bsnm{Hammoud}, \binits{B.}},
\bauthor{\bsnm{Khadiv}, \binits{M.}},
\bauthor{\bsnm{Bogdanovic}, \binits{M.}},
\bauthor{\bsnm{Berenz}, \binits{V.}},
\bauthor{\bsnm{Viereck}, \binits{J.}}, \betal:
\bctitle{Trifinger: An open-source robot for learning dexterity}.
In: \bbtitle{Conference on Robot Learning},
vol. \bseriesno{155},
pp. \bfpage{1871}--\blpage{1882}
(\byear{2021}).
\bcomment{PMLR}
\end{bchapter}
\endbibitem

\bibitem[\protect\citeauthoryear{Sutton and Barto}{2018}]{sutton2018reinforcement}
\begin{botherref}
\oauthor{\bsnm{Sutton}, \binits{R.S.}},
\oauthor{\bsnm{Barto}, \binits{A.G.}}:
Reinforcement learning: An introduction.
MIT press
(2018)
\end{botherref}
\endbibitem

\bibitem[\protect\citeauthoryear{Rissanen}{1978}]{rissanen1978modeling}
\begin{barticle}
\bauthor{\bsnm{Rissanen}, \binits{J.}}:
\batitle{Modeling by shortest data description}.
\bjtitle{Automatica}
\bvolume{14}(\bissue{5}),
\bfpage{465}--\blpage{471}
(\byear{1978})
\doiurl{10.1016/0005-1098(78)90005-5}
\end{barticle}
\endbibitem

\bibitem[\protect\citeauthoryear{Gr{\"u}nwald et~al.}{2007}]{grunwald2007minimum}
\begin{botherref}
\oauthor{\bsnm{Gr{\"u}nwald}, \binits{P.D.}}, et al.:
The minimum description length principle.
MIT Press Books
\textbf{1}
(2007)
\end{botherref}
\endbibitem

\bibitem[\protect\citeauthoryear{Rousseeuw}{1987}]{rousseeuw1987silhouettes}
\begin{barticle}
\bauthor{\bsnm{Rousseeuw}, \binits{P.J.}}:
\batitle{Silhouettes: A graphical aid to the interpretation and validation of cluster analysis}.
\bjtitle{Journal of Computational and Applied Mathematics}
\bvolume{20},
\bfpage{53}--\blpage{65}
(\byear{1987})
\doiurl{10.1016/0377-0427(87)90125-7}
\end{barticle}
\endbibitem

\bibitem[\protect\citeauthoryear{Cuturi and Blondel}{2017}]{cuturi2017soft}
\begin{bchapter}
\bauthor{\bsnm{Cuturi}, \binits{M.}},
\bauthor{\bsnm{Blondel}, \binits{M.}}:
\bctitle{Soft-{DTW}: a differentiable loss function for time-series}.
In: \bbtitle{International Conference on Machine Learning},
vol. \bseriesno{70},
pp. \bfpage{894}--\blpage{903}
(\byear{2017}).
\bcomment{PMLR}
\end{bchapter}
\endbibitem

\bibitem[\protect\citeauthoryear{Bagirov et~al.}{2023}]{bagirov2023finding}
\begin{barticle}
\bauthor{\bsnm{Bagirov}, \binits{A.M.}},
\bauthor{\bsnm{Aliguliyev}, \binits{R.M.}},
\bauthor{\bsnm{Sultanova}, \binits{N.}}:
\batitle{Finding compact and well-separated clusters: Clustering using silhouette coefficients}.
\bjtitle{Pattern Recognition}
\bvolume{135},
\bfpage{109144}
(\byear{2023})
\doiurl{10.1016/j.patcog.2022.109144}
\end{barticle}
\endbibitem

\bibitem[\protect\citeauthoryear{Punhani et~al.}{2022}]{punhani2022binning}
\begin{barticle}
\bauthor{\bsnm{Punhani}, \binits{A.}},
\bauthor{\bsnm{Faujdar}, \binits{N.}},
\bauthor{\bsnm{Mishra}, \binits{K.K.}},
\bauthor{\bsnm{Subramanian}, \binits{M.}}:
\batitle{Binning-based silhouette approach to find the optimal cluster using {K}-means}.
\bjtitle{IEEE Access}
\bvolume{10},
\bfpage{115025}--\blpage{115032}
(\byear{2022})
\doiurl{10.1109/ACCESS.2022.3215568}
\end{barticle}
\endbibitem

\bibitem[\protect\citeauthoryear{Shutaywi and Kachouie}{2021}]{shutaywi2021silhouette}
\begin{botherref}
\oauthor{\bsnm{Shutaywi}, \binits{M.}},
\oauthor{\bsnm{Kachouie}, \binits{N.N.}}:
Silhouette analysis for performance evaluation in machine learning with applications to clustering.
Entropy
\textbf{23}(6)
(2021)
\doiurl{10.3390/e23060759}
\end{botherref}
\endbibitem

\bibitem[\protect\citeauthoryear{Henderson et~al.}{2018}]{henderson2018deep}
\begin{botherref}
\oauthor{\bsnm{Henderson}, \binits{P.}},
\oauthor{\bsnm{Islam}, \binits{R.}},
\oauthor{\bsnm{Bachman}, \binits{P.}},
\oauthor{\bsnm{Pineau}, \binits{J.}},
\oauthor{\bsnm{Precup}, \binits{D.}},
\oauthor{\bsnm{Meger}, \binits{D.}}:
Deep reinforcement learning that matters.
Proceedings of the AAAI Conference on Artificial Intelligence
\textbf{32}(1)
(2018)
\doiurl{10.1609/aaai.v32i1.11694}
\end{botherref}
\endbibitem

\bibitem[\protect\citeauthoryear{Eimer et~al.}{2023}]{eimer2023hyperparameters}
\begin{bchapter}
\bauthor{\bsnm{Eimer}, \binits{T.}},
\bauthor{\bsnm{Lindauer}, \binits{M.}},
\bauthor{\bsnm{Raileanu}, \binits{R.}}:
\bctitle{Hyperparameters in reinforcement learning and how to tune them}.
In: \bbtitle{International Conference on Machine Learning},
vol. \bseriesno{202},
pp. \bfpage{9104}--\blpage{9149}
(\byear{2023}).
\bcomment{PMLR}
\end{bchapter}
\endbibitem

\bibitem[\protect\citeauthoryear{Kiran and Ozyildirim}{2022}]{kiran2022hyperparameter}
\begin{botherref}
\oauthor{\bsnm{Kiran}, \binits{M.}},
\oauthor{\bsnm{Ozyildirim}, \binits{M.}}:
Hyperparameter tuning for deep reinforcement learning applications.
arXiv preprint arXiv:2201.11182
(2022)
\end{botherref}
\endbibitem

\bibitem[\protect\citeauthoryear{Peters et~al.}{2011}]{peters2011causal}
\begin{barticle}
\bauthor{\bsnm{Peters}, \binits{J.}},
\bauthor{\bsnm{Janzing}, \binits{D.}},
\bauthor{\bsnm{Scholkopf}, \binits{B.}}:
\batitle{Causal inference on discrete data using additive noise models}.
\bjtitle{IEEE Transactions on Pattern Analysis and Machine Intelligence}
\bvolume{33}(\bissue{12}),
\bfpage{2436}--\blpage{2450}
(\byear{2011})
\doiurl{10.1109/TPAMI.2011.71}
\end{barticle}
\endbibitem

\bibitem[\protect\citeauthoryear{Vowels et~al.}{2022}]{vowels2022d}
\begin{botherref}
\oauthor{\bsnm{Vowels}, \binits{M.J.}},
\oauthor{\bsnm{Camgoz}, \binits{N.C.}},
\oauthor{\bsnm{Bowden}, \binits{R.}}:
D’ya like {DAG}s? a survey on structure learning and causal discovery.
ACM Comput. Surv.
\textbf{55}(4)
(2022)
\doiurl{10.1145/3527154}
\end{botherref}
\endbibitem

\bibitem[\protect\citeauthoryear{Glymour et~al.}{2019}]{glymour2019review}
\begin{botherref}
\oauthor{\bsnm{Glymour}, \binits{C.}},
\oauthor{\bsnm{Zhang}, \binits{K.}},
\oauthor{\bsnm{Spirtes}, \binits{P.}}:
Review of causal discovery methods based on graphical models.
Frontiers in Genetics
\textbf{10}
(2019)
\doiurl{10.3389/fgene.2019.00524}
\end{botherref}
\endbibitem

\bibitem[\protect\citeauthoryear{Hasan et~al.}{2023}]{hasan2023survey}
\begin{botherref}
\oauthor{\bsnm{Hasan}, \binits{U.}},
\oauthor{\bsnm{Hossain}, \binits{E.}},
\oauthor{\bsnm{Gani}, \binits{M.O.}}:
A survey on causal discovery methods for temporal and non-temporal data.
arXiv preprint arXiv:2303.15027
(2023)
\end{botherref}
\endbibitem

\end{thebibliography}

\end{document}